\documentclass[journal]{IEEEtran}

\usepackage{amssymb}
\usepackage{latexsym}
\usepackage{lineno,hyperref}
\usepackage{amsmath}
\usepackage{amsfonts}
\usepackage{tabularx}
\usepackage{hhline}
\usepackage{xcolor}
\usepackage{colortbl}
\usepackage{subcaption}
\usepackage{multirow}
\usepackage{wrapfig}
\usepackage{graphicx}
\usepackage[export]{adjustbox}[2011/08/13]
\usepackage[justification=centering]{caption}
\modulolinenumbers[5]

\newcommand{\sze}{0.12}
\newcommand{\hor}{5pt}
\newcommand{\ver}{7pt}
\begin{document}

\title{Attention! A Lightweight 2D Hand Pose \\Estimation Approach }

\author{Nicholas Santavas, \IEEEmembership{Student Member, IEEE}, Ioannis Kansizoglou, Loukas Bampis, Evangelos Karakasis \\and Antonios Gasteratos, \IEEEmembership{Senior Member, IEEE} 
	
	\thanks{Nicholas Santavas, Ioannis Kansizoglou, Loukas Bampis, Evangelos Karakasis and Antonios Gasteratos are with Democritus University of Thrace, Department of Production and Management Engineering, Laboratory of Robotics and Automation, Xanthi, Greece. (e-mail: \{nsantava, ikansizo, lbampis, ekaraka, agaster\}@pme.duth.gr).}
}

\markboth{Journal of \LaTeX\ Class Files, Vol. X, No. X, XX XXXX}
{Shell \MakeLowercase{\textit{et al.}}: Bare Demo of IEEEtran.cls for IEEE Journals}
\maketitle
\maketitle

\begin{abstract}
 Vision based human pose estimation is an non-invasive technology for Human-Computer Interaction (HCI). The direct use of the hand as an input device provides an attractive interaction method, with no need for specialized sensing equipment, such as exoskeletons, gloves etc, but a camera. Traditionally, HCI is employed in various applications spreading in areas including manufacturing, surgery, entertainment industry and architecture, to mention a few. Deployment of vision based human pose estimation algorithms can give a breath of innovation to these applications. In this paper, we present a novel Convolutional Neural Network architecture, reinforced with a Self-Attention module. Our proposed model can be deployed on an embedded system due to its lightweight nature with just \textit{1.9 Million} parameters. The source code and qualitative results are publicly available \footnote{nsantavas.github.io}.
\end{abstract}


\IEEEpeerreviewmaketitle

\section{Introduction}

\IEEEPARstart{T}{he} role of computers and robots in our modern society keeps expanding and diversifying. As humanity breaks the technological barriers of the past, daily activities become more and more assisted by an excessive amount of operations based on the interaction between humans and computers. The development of more sophisticated systems frequently drives to complicated interaction ways that incommode their usage.

In order to democratize decision-making machines, straightforward ways of interaction need to be developed which imitate the relationship between humans \cite{zlotowski2015anthropomorphism}. A convenient way for a human to interact with machines can be achieved by means of natural dialogue. Such examples that are already implemented on virtual assistants are known as interactive conversational systems \cite{kepuska2018next}. Computer vision techniques are often applied in this field, e.g., for face and emotion recognition \cite{kansizoglou2019active}\cite{efremova2019face}, 3D face mesh representation for augmented reality applications \cite{kartynnik2019real}, action recognition \cite{vemulapalli2014human}, and finally, human body pose estimation \cite{cao2018openpose}\cite{zhang2019fast} and hand pose estimation \cite{chen2019pose}\cite{yuan2018depth}.

Human hand pose estimation is a long standing problem in the computer vision and graphics research fields, with a plethora of applications such as machine control, or augmented and virtual reality \cite{jang20153d}\cite{piumsomboon2013user}\cite{fang2015robotic}. Due to its importance, numerous solutions have been proposed in the related literature, with one of the most common being based on accurate 2D keypoint localization \cite{rehg1994visual}.

\begin{figure*}
	\centering
	\includegraphics[totalheight=0.22\textwidth ]{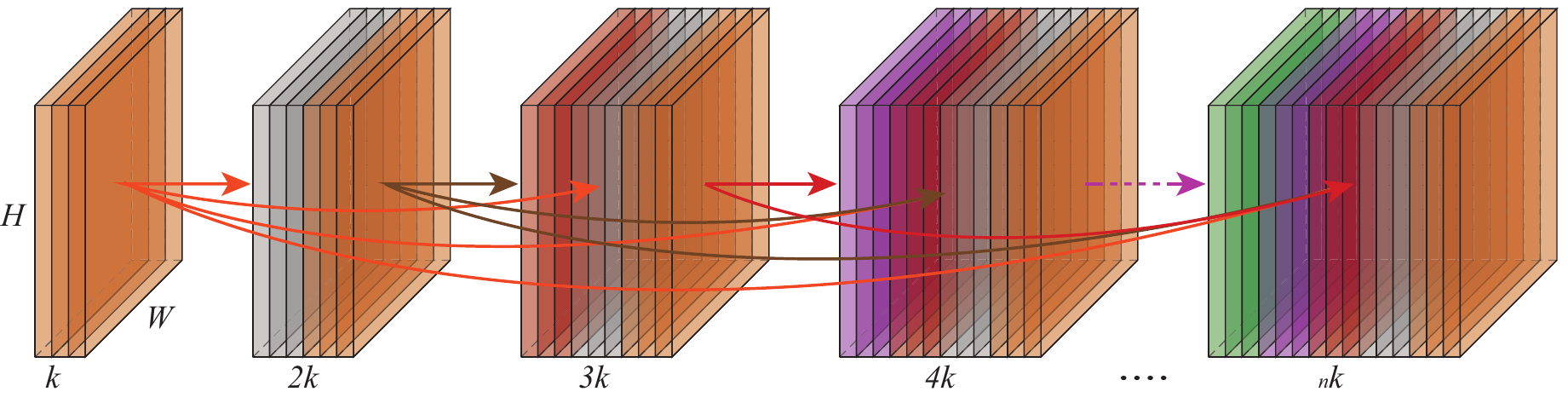}
	\caption{Dense Block with growth rate \textit{k} \cite{huang2017densely}.}
	\captionsetup{justification=centering}
	\label{fig:denseblock}
\end{figure*}

Despite the recent advances in the field of deep neural networks, this topic is still considered to be a challenging problem, that remains to be completely solved. Properties such as the hand's morphology, occlusions due to interaction with objects, appearance diversity due to clothing and jewelry, varying lightning conditions and different backgrounds, add extra burden to the nature of the problem. Nevertheless, unlike the human's body or face, hands have almost uniform shape and lack local characteristics. In addition, because of the few ten joints of a hand, there are myriads of different arbitrary poses. Thus, it becomes critical to localize more than 20 keypoints in each hand \cite{boukhayma20193d}\cite{iqbal2018hand} in order to accurately estimate its pose and use it as an input device.

In the paper at hand, we propose a computationally inexpensive CNN architecture for 2D direct hand pose estimation. Our contribution lies on the following novelties: 

\begin{itemize}
	
	\item The presentation of a single-stage end-to-end 2D hand pose estimation CNN architecture, which directly regresses the coordinates of the hand's keypoints from a single RGB image, without depending on the traditional two-stage pipeline\cite{newell2016stacked}\cite{wei2016convolutional}.
	
	\item The introduction of a novel and notably efficient block, dubbed as \textit{Attention Augmented Inverted Bottleneck Block}, the performance of which is thoroughly assessed.
	
	\item The design of a remarkably lightweight architecture based on the proposed block, robust to input shifts.
	
	\item We show that the exploitation of a self-attention mechanism \cite{bello2019attention}, combined with traditional convolutional layers, outperforms other computationally demanding state-of-the-art methods.
	
\end{itemize}
We evaluate our approach using various contemporal challenging datasets, which include images in-the-wild, occlusions and hand-object interactions, and compare the achieved performance to other state-of-the-art methods. 

The paper is organized as follows. Section \ref{Related Work} reviews of related works, Section \ref{Method} gives an aspect of our method, Section \ref{Evaluation} presents the experimental results and finally, in Section \ref{Conclusions} conclusions are drawn.

\section{Related Work}\label{Related Work}

Vision-based hand pose estimation has recently made significant progress. A vast amount of approaches uses Convolutional Neural Networks (CNNs) as a basis, due to their profound capability of extracting features from a given input. CNNs successfully perform 2D body pose estimation by classifying whether or not a body's joint is present in each pixel \cite{newell2016stacked}\cite{wei2016convolutional}. 
The proposed methods, also known as Convolutional Pose Machines (CPM), enforce a CNN to generate a set of heat maps, each of which, is expected to have its maximum activation value in the pixel that contains the corresponding keypoint. However, to refine the outcome, this procedure is applied iteratively upon the generated heat maps. Furthermore, the majority of hand pose estimation methods are also based upon the same approach \cite{boukhayma20193d}\cite{iqbal2018hand}, which leads to computationally expensive networks and complicated system architectures.\par
Another line of work aims to directly map the input image to the keypoints' coordinates on the plane or to a specific frame of reference for 2D and 3D pose estimation, respectively, known as holistic regression \cite{tekin2016direct}\cite{li20143d}. The abovementioned approach does not have to generate intermediate representations (pixel-wise classification) while also preserving the ability to understand global constraints and correlations between keypoints. However, it is claimed that holistic regression is not able to generalize and that translational variance diminish the predicted results \cite{wan2018dense}. Despite those conservations, the proposed work is based on such a technique, proving its capabilities when combined with a proper anti-aliasing filter and a robust feature extractor.

\section{Method}\label{Method}
In this section, we describe the structure of the proposed architecture and the key ingredients for estimating a hand's 2D keypoints, coordinates, given a single RGB image. Towards a solution for this challenge, we make use of a feed-forward CNN architecture that directly produces the coordinates in a single stage, without intermediate supervision. The network's architecture comprises two parts, the \textit{stem} and the rest, for now on dubbed as \textit{tail}

\subsection{Network's architecture}
The presented architecture is based on the successful idea of \textit{DenseNets} \cite{huang2017densely}. In a \textit{DenseNet}, each layer obtains additional inputs from all preceding ones and propagates its own feature-maps to all subsequent layers by a channel-wise concatenation, as shown in \figurename{ \ref{fig:denseblock}}. In such a way, this structure receives a ``collective knowledge'' from all previous layers.

To keep the total number of parameters as low as possible, we were inspired by a popular building unit, viz. the \textit{Inverted residual block}, which is a highly efficient feature extractor, designed especially for mobile use \cite{sandler2018mobilenetv2}. The replacement of the standard convolutional layer by depthwise separable ones offers a computation reduction by a factor:

\begin{equation}
k_{f}^{2}\cdot d_{o}/(k^{2}+d_{o}),
\end{equation}
where ${k_{f}}$ equals the kernel's size, and \textit{$d_{o}$} equals the output depth size. The first convolutional layer expands the depth size by an \textit{e} factor while the last squeezes it by dividing the input's depth size by the same factor. Here, $\textit{e}= 4$.

\subsubsection{Stem}
For \textit{stem}, we use a number of \textit{dense blocks} which, unlike the original design, it contains an \textit{inverted residual block}. According to  \cite{huang2017densely}, architectures with concatenated skip-connections maintain more information since they allow subsequent layers to reuse intermediate representations, which in turn, leads to increased performance.

A significant difference from the original block  regarding its non-linearity is that we use the lately proposed \textit{Mish} activation function \cite{misra2019mish}. \textit{Mish}, unlike ReLU, is a smooth non-monotonic activation function which is defined as:
\begin{equation}
f(x)=x \cdot \tanh(\ln(1+e^{x})),
\end{equation}
As mentioned in \cite{misra2019mish}, \textit{Mish} demonstrates better results than both
Swish \cite{ramachandran2017swish} and ReLU for classification tasks. After extensive experimentation with both Swish and ReLU, we confirmed the above behaviour for the regression   task at hand.

\subsubsection{Blur Pooling}
As it is widely known, many modern CNNs perform some sort of downsampling. A common practice for sub-sampling feature maps between convolutional layers is using either a pooling operation, or a strided convolution. In \cite{simoncelli1992shiftable}, it was explicitly discussed that a system based on both operations of convolution and sub-sampling lacks translation invariance, unless the translation is a multiple of each of the sub-sampling factors. Otherwise, sub-sampling creates alias that undermines the output. This property affects CNNs as well, since small spatial image transformations can lead to significant accuracy degradation \cite{engstrom2019exploring}\cite{azulay2018deep}. As stated in \cite{zhang2019making}, a feature extractor function $F \in \mathbb{R}^{H\times W \times C}$ is shift-equivariant when shifting the input equally shifts the output, making shifting and feature extraction commutable:
\begin{equation}
Shift_{\Delta h,\Delta w}(F(X)) = F(Shift_{\Delta h,\Delta w}(X))\qquad \forall(\Delta h, \Delta w).
\end{equation}
Furthermore, a representation is shift-invariant if shifting the inputs results in an identical representation:
\begin{equation}
F(X) = F(Shift_{\Delta h,\Delta w}(X))\qquad  \forall(\Delta h, \Delta w).
\end{equation}

Regural pooling methods break shift-equivariance. To overcome this issue, we propose the adaptation of an anti-aliasing filter, which is convolved with feature maps \cite{zhang2019making}, with stride 2 to reduce spatial resolution. The method provides the ability to choose between different size of kernels, producible by a box filter. The following implements the anti-aliasing filter $Filt$.
\begin{equation}
B_{n}[x] =\left\{
\begin{array}{ll}
1,  & \mbox{for } 0 \leq x < n, \\
0, & \mbox{ } otherwise,
\end{array}
\right.
\end{equation}
\begin{equation}
Box_{m} = B_{n}*B_{n},
\end{equation}
\begin{equation}
Filt = Box_{m} \otimes Box_{m},
\end{equation}
where $\otimes$ denotes the outer product,  $n \in \mathbb{N}^{*}$, $x \in \mathbb{Z}$ and \mbox{$m= 2n-1$}.
In our case, the utilized anti-aliasing filter $Filt$ $n=2$.

\subsubsection{Attention Augmented Inverted Bottleneck Block}

Attention mechanisms enable a neural network to focus more on relevant elements of the input than on irrelevant parts. Visual attention is one of the most influential ideas in the deep learning research field. Attention mechanisms and especially self-attention, are powerful building blocks for processing not only text but also images. Many visual attention mechanisms have been proposed to enhance the convolutions' already proved performance \cite{vaswani2017attention}\cite{bello2019attention}.

\begin{figure}
	\centering
	\includegraphics[totalheight=0.7\linewidth ]{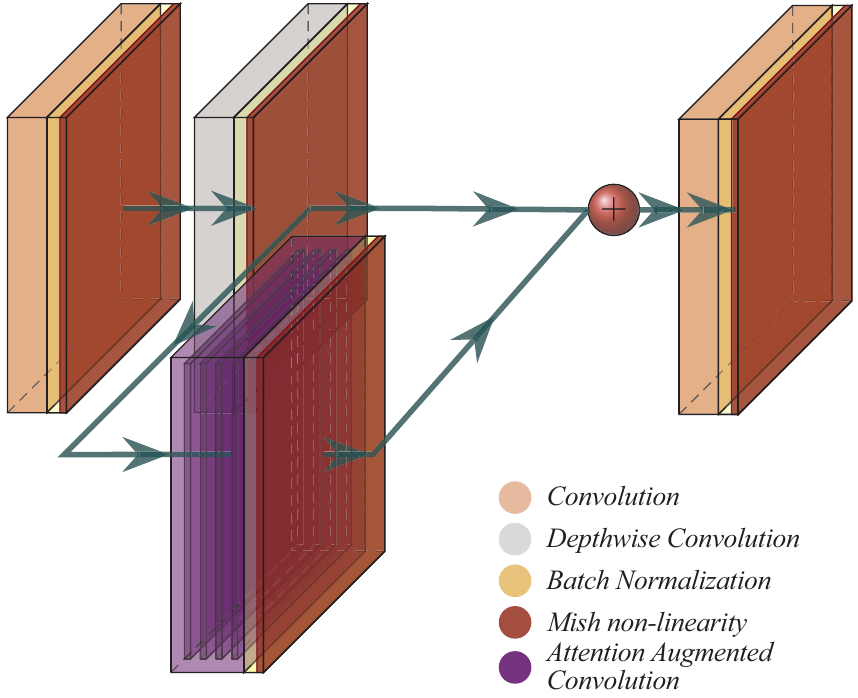}
	\caption{Our proposed Attention Augmented Inverted Bottleneck Layer.}
	\label{fig:aaibl}
\end{figure}

The general idea is that given a query and a set of key elements, the attention mechanism aggregates w.r.t the trainable parameters, the resemblance between key-query pairs. Multiple attention functions provide the ability to attend multiple representation subspaces and spatial positions. Finally, each head's output is linearly aggregated with learnable weights \cite{zhu2019empirical}. Our work was inspired by a design proposed in \cite{zhu2019empirical}, in which a self-attention mechanism enfolds a standard residual block. More specifically, we implement an Attention Augmented Convolutional layer \cite{bello2019attention}, which embeds an \textit{inverted bottleneck block}, by adding its output to the product of the Depthwise Separable Convolutional layer, as shown in \figurename{ \ref{fig:aaibl}}. 

A self-attention mechanism achieves better results when combined with convolutional layers \cite{bello2019attention}. In practice, a self-attention module uses three sets of learnable parameters $W^{Q}, W^{K}, W^{V},$ where $Q,K,V$ stand for \textit{Query}, \textit{Key} and \textit{Value}, respectively. According to \cite{vaswani2017attention}, an input tensor $ T \in \mathbb{R}^{H\times W \times F_{in}} $, is flattened to a matrix $X \in \mathbb{R}^{HW\times F_{in}}$ and then forwarded to the Transformer attention architecture. Since it has been found beneficial to apply self-attention multiple times, Eq. \ref{eq:1} is applied once for each attention head,  producing $O_{[1,.,h]}$ outputs, where $ h\in \mathbb{N}^{*}$.

\begin{equation}\label{eq:1}
O_{h}=Softmax\left(\frac{(XW^{Q})(XW^{K})^{T}}{\sqrt{d_{k}^{h}}}\right)(XW^{V}).
\end{equation}
$W^{Q}, W^{K} \in \mathbb{R}^{F_{in} \times d_{k}^{h}}$ and $W^{V} \in \mathbb{R}^{F_{in}\times d_{v}^{h}}$. 
The output of each head is then concatenated with the remaining, forming the Multihead Attention mechanism.
\begin{equation}
MHA(X)=Concat\left[ O_{1},..,O_{h}\right] W^{O},
\end{equation}\\
where $W^{O} \in \mathbb{R}^{d_{v}\times d_{x}}$ is a trainable matrix which linearly transforms the aggregated output of each head. We refer to the \textit{Values'} depth as $d_{v}$, \textit{Queries'} depth as $d_{k}$ and the number of heads as $N_{h}$.
\begin{table}
	\renewcommand*{\arraystretch}{1.4}
	\caption{{{Network's Architecture}. The growth rate is \textit{k}=10}.}
	\label{table:arch}
	\centering
	\resizebox{\linewidth}{!}{
		\begin{tabular}{ccc}
			\hline
			\textbf{Layers}                      & \textbf{Output Size} & \textbf{Architecture}                                                         \\ 
			\hline

			\multicolumn{1}{l}{Dense Block (1)}  & 224$\times$224            & {[}Inverted bottleneck layer] $\times$8                                               \\
			\multicolumn{1}{l}{Transition Layer} & 112 $\times$ 112            & \begin{tabular}[c]{@{}c@{}}1 $\times$ 1 conv $\times$ 64\\3 $\times$ 3 BlurPool, s2\end{tabular}   \\
			\multicolumn{1}{l}{Dense Block (2)~} & 112 $\times$ 112            & {[}Inverted bottleneck layer] $\times$8                                              \\
			\multicolumn{1}{l}{Transition Layer} & 56 $\times$ 56              & \begin{tabular}[c]{@{}c@{}}1 $\times$ 1 conv $\times$ 64\\3 $\times$ 3 BlurPool, s2\end{tabular}   \\
			Dense Block (3)                      & 56 $\times$ 56              & {[}Attention Augmented Inverted bottleneck layer] $\times$6                          \\
			Transition Layer                     & 28 $\times$ 28              & \begin{tabular}[c]{@{}c@{}}1 $\times$ 1 conv $\times$ 64\\3 $\times$ 3 BlurPool, s2\end{tabular}   \\
			Dense Block (4)                      & 28 $\times$ 28              & {[}Attention Augmented Inverted bottleneck layer] $\times$8                          \\
			Transition Layer                     & 14 $\times$ 14              & \begin{tabular}[c]{@{}c@{}}1 $\times$ 1 conv $\times$ 64\\3 $\times$ 3 BlurPool, s2\end{tabular}   \\
			Dense Block (5)                      & 14 $\times$ 14              & {[}Attention Augmented Inverted bottleneck layer] $\times$10                         \\
			Transition Layer                     & 7 $\times$ 7                & \begin{tabular}[c]{@{}c@{}}1 $\times$ 1 conv $\times$ 64\\3 $\times$ 3 BlurPool, s2\end{tabular}   \\
			Dense Block (6)                      & 7 $\times$ 7                & {[}Attention Augmented Inverted bottleneck layer] $\times$12                         \\
			Transition Layer                     & 4 $\times$ 4                & \begin{tabular}[c]{@{}c@{}}1 $\times$ 1 conv $\times$ 128\\3 $\times$ 3 BlurPool, s2\end{tabular}  \\
			Dense Block (7)                      & 4 $\times$ 4                & {[}Attention Augmented Inverted bottleneck layer] $\times$14                         \\
			Transition Layer                     & 2 $\times$ 2                & \begin{tabular}[c]{@{}c@{}}1 $\times$ 1 conv $\times$ 128\\3 $\times$ 3 BlurPool, s2\end{tabular}  \\
			Dense Block (8)                      & 2 $\times$ 2                & {[}Attention Augmented Inverted bottleneck layer] $\times$32                         \\
			AA-Bottleneck                        & 2 $\times$ 2                & {[}Attention Augmented Inverted bottleneck layer] $\times$1                          \\
			& 1 $\times$ 1                & 2 $\times$ 2 Average Pooling, s2                                                     \\
			& 1 $\times$ 1 $\times$ 42           & 1 $\times$ 1 conv $\times$ 42                                                    \\ 
			\hline
	\end{tabular}}
\end{table}
\begin{figure}[!t]
	\centering
	\begin{subfigure}{0.4\textwidth}
		\includegraphics[width=\textwidth]{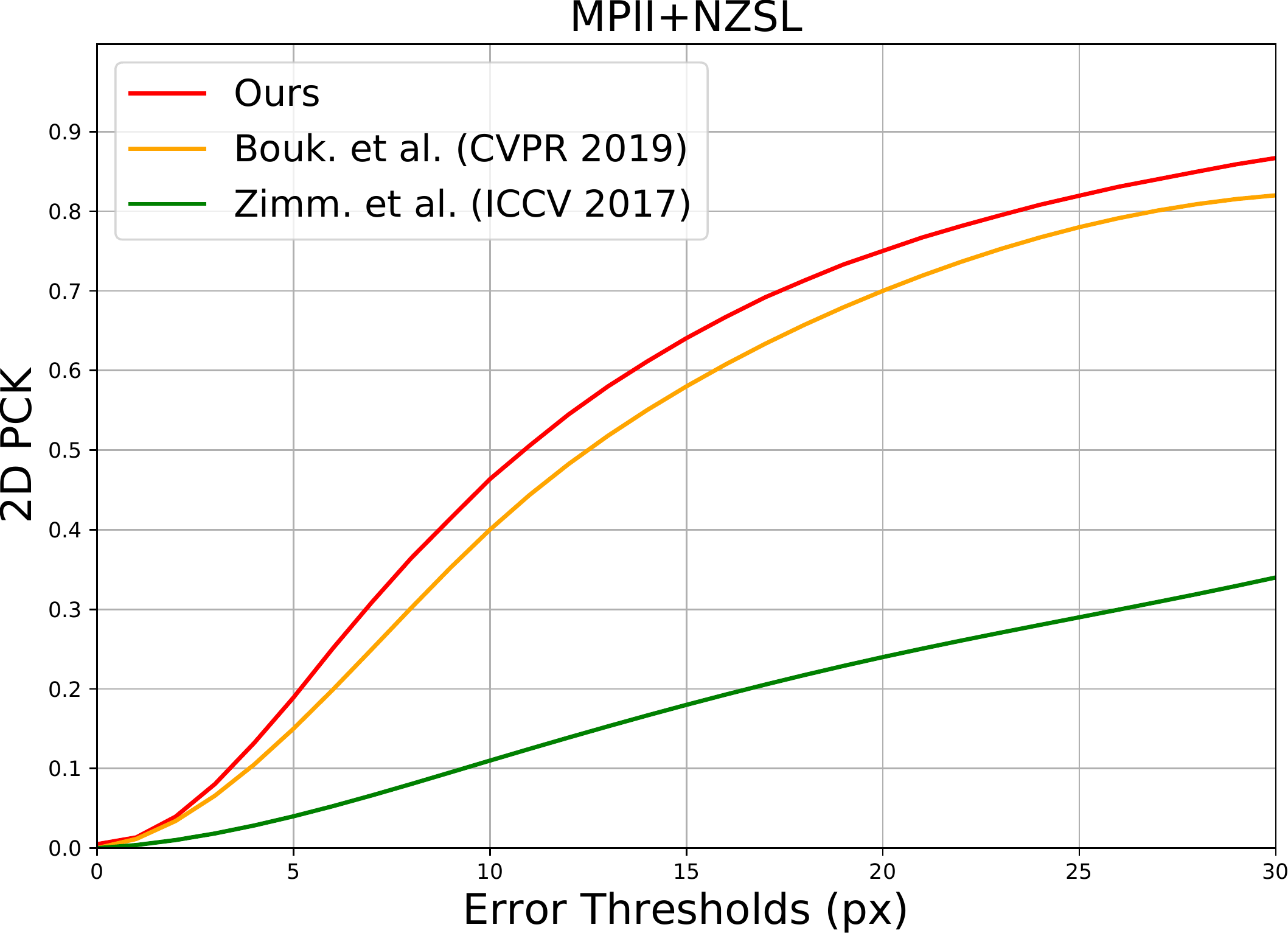}
		\caption{{ PCK curves on MPII+NZSL testing set}.}
		\label{grapha}
	\end{subfigure}
	\begin{subfigure}{0.4\textwidth}
		\includegraphics[width=\textwidth]{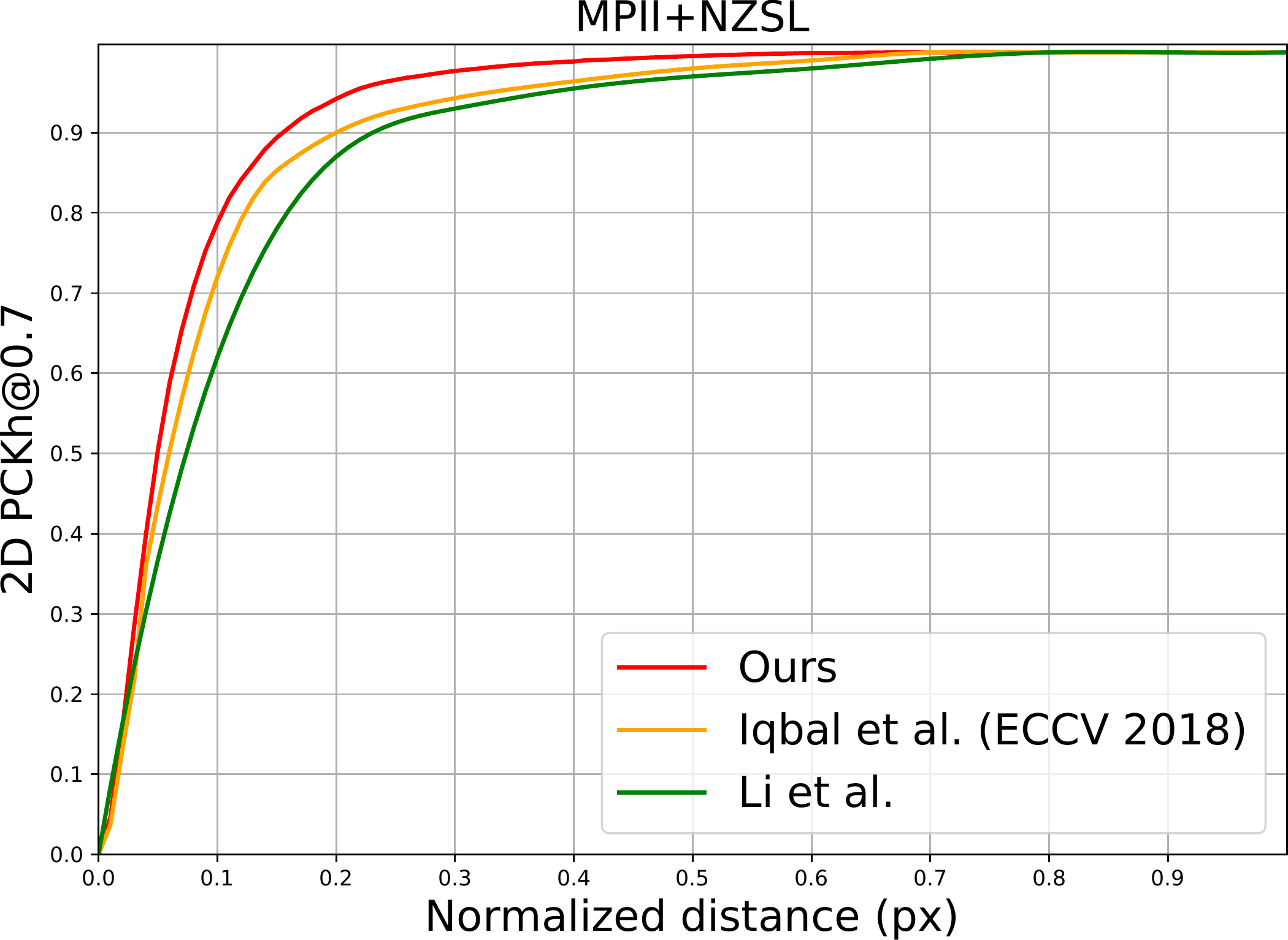}
		\caption{{ PCKh curves on MPII+NZSL testing set}.}
		\label{graphb}
	\end{subfigure}
	
	\begin{subfigure}{0.4\textwidth}
		\includegraphics[width=\textwidth]{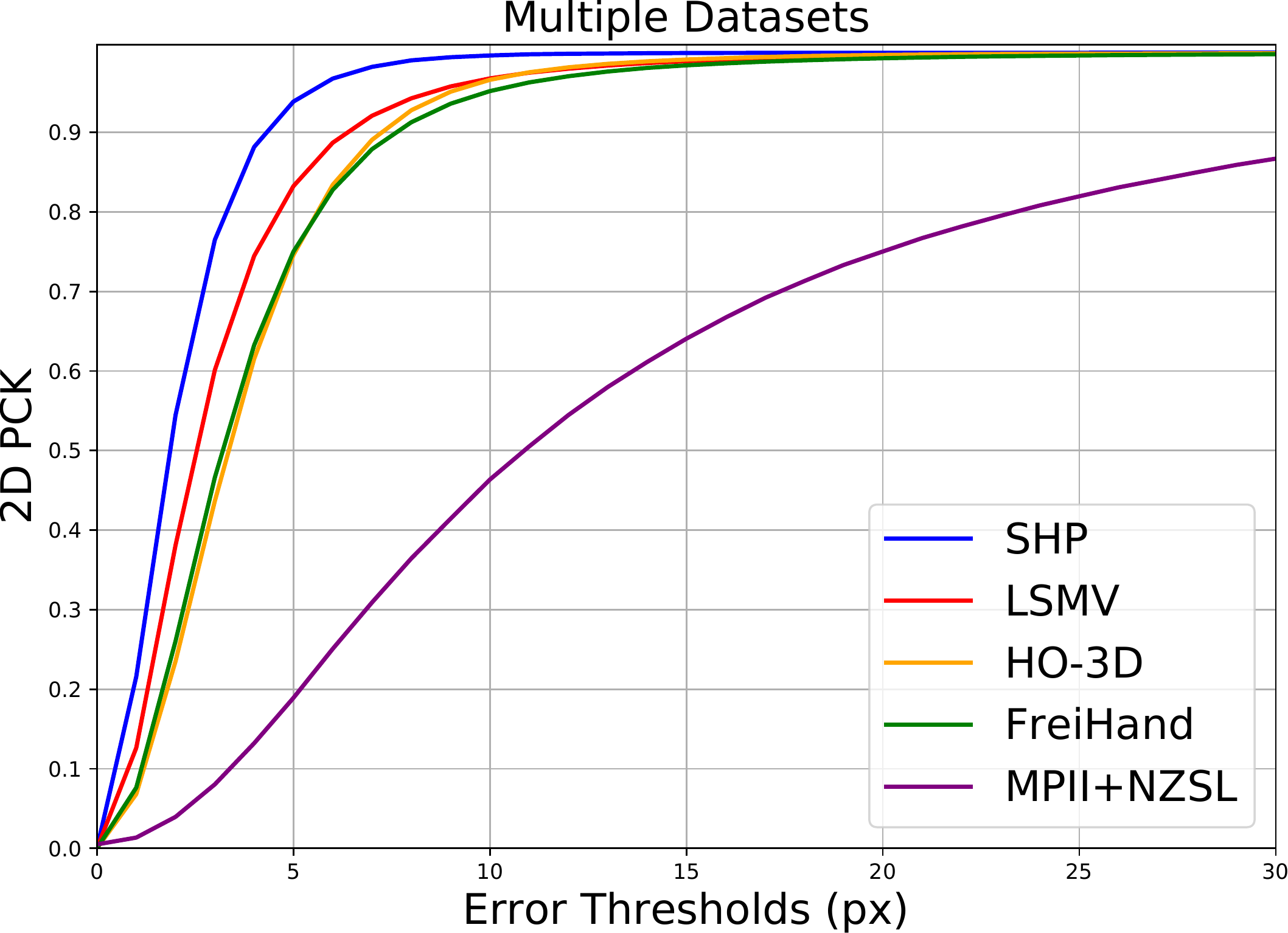}
		\caption{{ PCK curves of our method on different datasets}.}
		\label{graphc}
	\end{subfigure}
	
	\caption{Performance Evaluation.}
	\label{evaluation}
\end{figure}
\begin{figure*}[!th]
	
	\centering
	
	\begin{subfigure}[c]{\sze\linewidth}
		\includegraphics[width=\linewidth]{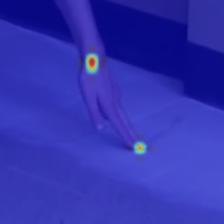}
	\end{subfigure}
	\hspace{\hor}
	\begin{subfigure}[c]{\sze\linewidth}
		\includegraphics[width=\linewidth]{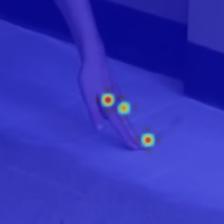}
	\end{subfigure}
	\hspace{\hor}
	\begin{subfigure}[c]{\sze\linewidth}
		\includegraphics[width=\linewidth]{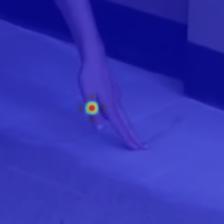}
	\end{subfigure}
	\hspace{\hor}
	\begin{subfigure}[c]{\sze\linewidth}
		\includegraphics[width=\linewidth]{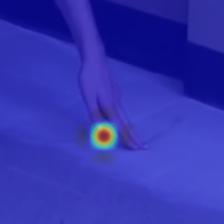}
	\end{subfigure}
	\hspace{\hor}
	\begin{subfigure}[c]{\sze\linewidth}
		\includegraphics[width=\linewidth]{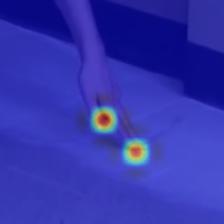}
	\end{subfigure}
	\hspace{\hor}
	\begin{subfigure}[c]{\sze\linewidth}
		\includegraphics[width=\linewidth]{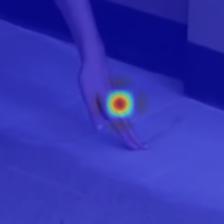}
	\end{subfigure}
	
	\vspace{\ver}	
	\begin{subfigure}[c]{\sze\linewidth}
		\includegraphics[width=\linewidth]{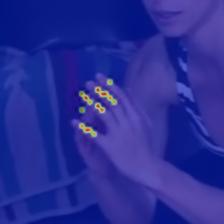}
	\end{subfigure}
	\hspace{\hor}
	\begin{subfigure}[c]{\sze\linewidth}
		\includegraphics[width=\linewidth]{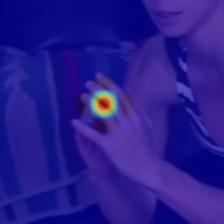}
	\end{subfigure}
	\hspace{\hor}
	\begin{subfigure}[c]{\sze\linewidth}
		\includegraphics[width=\linewidth]{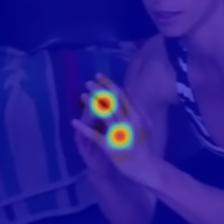}
	\end{subfigure}
	\hspace{\hor}
	\begin{subfigure}[c]{\sze\linewidth}
		\includegraphics[width=\linewidth]{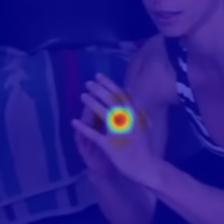}
	\end{subfigure}
	\hspace{\hor}
	\begin{subfigure}[c]{\sze\linewidth}
		\includegraphics[width=\linewidth]{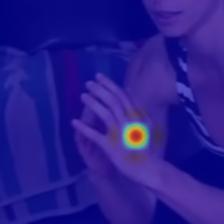}
	\end{subfigure}
	\hspace{\hor}
	\begin{subfigure}[c]{\sze\linewidth}
		\includegraphics[width=\linewidth]{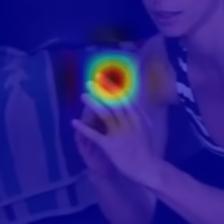}
	\end{subfigure}

	\vspace{\ver}	
	\begin{subfigure}[c]{\sze\linewidth}
		\includegraphics[width=\linewidth]{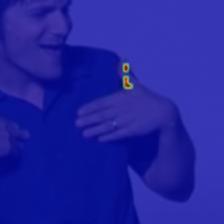}
	\end{subfigure}
	\hspace{\hor}
	\begin{subfigure}[c]{\sze\linewidth}
		\includegraphics[width=\linewidth]{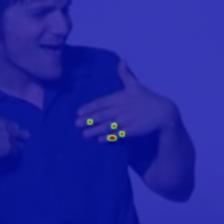}
	\end{subfigure}
	\hspace{\hor}
	\begin{subfigure}[c]{\sze\linewidth}
		\includegraphics[width=\linewidth]{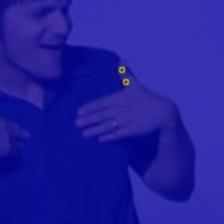}
	\end{subfigure}
	\hspace{\hor}
	\begin{subfigure}[c]{\sze\linewidth}
		\includegraphics[width=\linewidth]{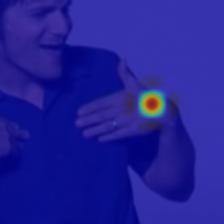}
	\end{subfigure}
	\hspace{\hor}
	\begin{subfigure}[c]{\sze\linewidth}
		\includegraphics[width=\linewidth]{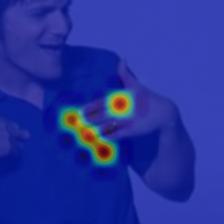}
	\end{subfigure}
	\hspace{\hor}
	\begin{subfigure}[c]{\sze\linewidth}
		\includegraphics[width=\linewidth]{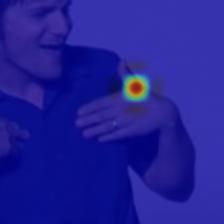}
	\end{subfigure}

	\vspace{\ver}	
	\begin{subfigure}[c]{\sze\linewidth}
		\includegraphics[width=\linewidth]{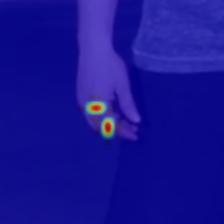}
	\end{subfigure}
	\hspace{\hor}
	\begin{subfigure}[c]{\sze\linewidth}
		\includegraphics[width=\linewidth]{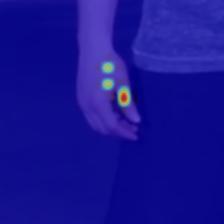}
	\end{subfigure}
	\hspace{\hor}
	\begin{subfigure}[c]{\sze\linewidth}
		\includegraphics[width=\linewidth]{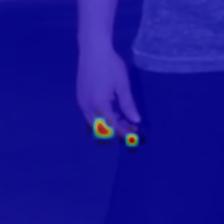}
	\end{subfigure}
	\hspace{\hor}
	\begin{subfigure}[c]{\sze\linewidth}
		\includegraphics[width=\linewidth]{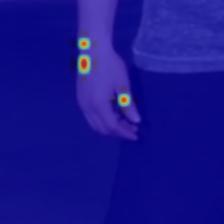}
	\end{subfigure}
	\hspace{\hor}
	\begin{subfigure}[c]{\sze\linewidth}
		\includegraphics[width=\linewidth]{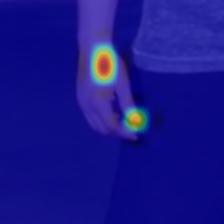}
	\end{subfigure}
	\hspace{\hor}
	\begin{subfigure}[c]{\sze\linewidth}
		\includegraphics[width=\linewidth]{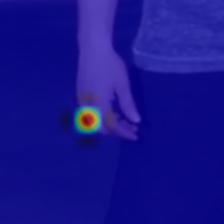}
	\end{subfigure}
	
	\vspace{\ver}	
	\begin{subfigure}[c]{\sze\linewidth}
		\includegraphics[width=\linewidth]{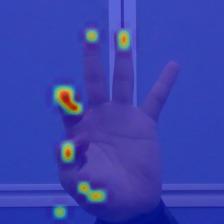}
	\end{subfigure}
	\hspace{\hor}
	\begin{subfigure}[c]{\sze\linewidth}
		\includegraphics[width=\linewidth]{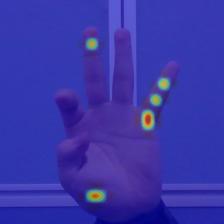}
	\end{subfigure}
	\hspace{\hor}
	\begin{subfigure}[c]{\sze\linewidth}
		\includegraphics[width=\linewidth]{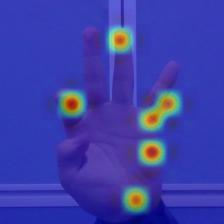}
	\end{subfigure}
	\hspace{\hor}
	\begin{subfigure}[c]{\sze\linewidth}
		\includegraphics[width=\linewidth]{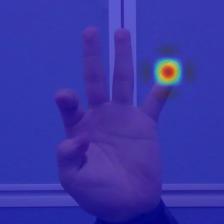}
	\end{subfigure}
	\hspace{\hor}
	\begin{subfigure}[c]{\sze\linewidth}
		\includegraphics[width=\linewidth]{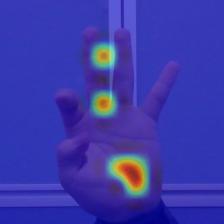}
	\end{subfigure}
	\hspace{\hor}
	\begin{subfigure}[c]{\sze\linewidth}
		\includegraphics[width=\linewidth]{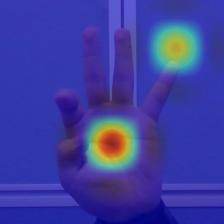}
	\end{subfigure}

	\caption{Representative feature maps of our proposed Attention Augmented Inverted Bottleneck Layer.}
	\label{heatmaps}
\end{figure*}

An inherent characteristic of self-attention is that it is equivariant to an input's reordering. This essentially means that any spatial information is not maintained, which is prohibitive for vision tasks due to the structured nature of the images. To alleviate the limitation, a trainable \textit{positional encoding} is assigned to each pixel of the image. The relative position of both width and height, between each \textit{Query} and \textit{Key} pixel, is represented by two matrices that contain a \textit{relative position embedding} for every pixel pair. The relationship's strength between two pixels $i,j$ is computed as:

\begin{equation}
l_{i,j}= \frac{q_{i}^{T}} {\sqrt{d_{k}^{h}}} (k_{j}+r_{j_{x}-i_{x}}^{W}+r_{j_{y}-i_{y}}^{H}),
\end{equation} 
where $q_{i}$ and $k_{j}$ are the \textit{Query} and \textit{Key} vectors for pixels $i,j$, while $r_{j_{x}-i_{x}}^{W}$ and $r_{j_{y}-i_{y}}^{H}$ are learned embeddings for relative width and height, respectively.

Each attention head enhanced by \textit{relative position embeddings} becomes:
\begin{equation}
O_{h}=Softmax\left(\frac{(XW^{Q})(XW^{K})^{T}+S^{rel}_{H}+S^{rel}_{W}}{\sqrt{d_{k}^{h}}}\right)(XW^{V}),
\end{equation} 
where $S^{rel}_{H}$, $S^{rel}_{H} \in \mathbb{R}^{HW\times HW}$ are matrices of \textit{relative positional embeddings} for every pixel pair.

As previously mentioned, this type of visual attention has the ability to attend feature subspaces and spatial positions simultaneously, both due to the attention mechanism that introduces additional feature maps and the convolution operator. The last part of the Attention's Augmented Convolution integration includes the concatenation between the convolutional operator and Multiheaded Attention's output.

\begin{equation}
AAC(X)=concat\left[ Conv(X),MHA(X)\right].
\end{equation}
Denoted as $ u=\frac{d_v}{F_{out}}$ is the ratio between attention depth size and the output depth size, while $\kappa = \frac{d_{k}}{F_{out}}$ is the ratio of key depth over the output depth.

For the network's \textit{tail}, we implement recurrently \textit{dense blocks} that contain the \textit{Attention Augmented Inverted Bottleneck layer}, with a similar manner proposed for the \textit{stem}. 

\subsubsection{Downsampling}
To downsample the feature maps between \textit{dense blocks}, a \textit{transition layer} is used, which comprises a pointwise convolutional layer for depth reduction, an Blur Pooling filter with stride 2 and finally, batch normalization. 

\subsection{Training}
During training, \textit{Cyclical Learning Rate} \cite{smith2017cyclical}  with triangular policy was used with \textit{Stochastic Gradient Descent} optimizer. The selected hyper-parameters are, $stepsize=6$, minimum learning rate of $10^{-4}$ and maximum learning rate of $10^{-1}$. The \textit{batchsize} equals to 256, and the training was executed using \textit{Tensor Processing Units} (TPUs) on the cloud, provided by Google. Finally, a mixed-precision training policy was used by exploiting both 16-bit (bfloat16) and 32-bit (float32) floating-point types \cite{micikevicius2017mixed}. This practice resulted to memory gain, which in turn led to greater batch size, smaller model size and faster execution time. Table \ref{table:arch} explicitly shows the model's architecture, totaling just \textit{1.9M} parameters and \textit{7.1 Million} FLOPs in terms of computational demands, which was developed using the TensorFlow library \cite{abadi2016tensorflow}.

\begin{table}[!b]
	\renewcommand*{\arraystretch}{1.3}
	\centering
	\small
	\caption{{Quantitative performance results}.}
	\label{table:results}
	\begin{tabular}{cccc} 
		\hline
		& \multirow{2}{*}{\textbf{AUC }} & \multicolumn{2}{c}{\textbf{EPE(px) }}  \\
		&                                & \textbf{Mean } & \textbf{Median }      \\ 
		\hline
		\multicolumn{4}{c}{MPII+NZSL Dataset}                                                              \\
		\hline
		Zimm. et al. (ICCV 2017) & 0.17                           & 59.4           & -                     \\
		Bouk. et al. (CVPR 2019) & 0.50                           & 18.95          & -                     \\
		\textbf{Ours }           & \textbf{0.55 }                 & \textbf{16.1 } & \textbf{11 }          \\
		\hline
		\multicolumn{4}{c}{LSMV Dataset}                                                                   \\
		\hline
		Gomez-Donoso et al.      & -                              & 10             & -                     \\
		Li et al.                & -                              & 8              & -                     \\
		\textbf{Ours }           & 0.89                           & \textbf{3.3 }  & 2.5                   \\
		\hline
		\multicolumn{4}{c}{Stereo Hand Pose Dataset}                                                       \\
		\hline
		Zimm et al. (ICCV 2017)  & 0.81                           & 5              & 5.5                   \\
		\textbf{Ours }           & \textbf{0.92 }                 & \textbf{2.2 }  & \textbf{1.8 }         \\
		\hline
		\multicolumn{4}{c}{HO-3D Dataset}                                                                  \\
		\hline
		Ours                     & 0.87                           & 3.9            & 3.3                   \\
		\hline
		\multicolumn{4}{c}{FreiHand Dataset}                                                               \\
		\hline
		Ours                     & 0.87                           & 4              & 3.1                   \\
		\hline
	\end{tabular}
\end{table}

\begin{table*}[!h]
	\renewcommand*{\arraystretch}{1.3}
	\centering
	\caption{{Different architectures utilized in the ablation study}.}
	\label{table:diff_archs}
	\begin{tabular}{ccccccccccccc}
		\hline
		& \textbf{Arch 1} & \textbf{Arch 2} & \textbf{Arch 3} & \textbf{Arch 4} & \textbf{Arch 5} & \textbf{Arch 6} & \textbf{Arch 7} & \textbf{Arch 8} & \textbf{Arch 9} & \textbf{Arch 10} & \textbf{Arch 11} & \textbf{Arch 12} \\
		\hline
		\textbf{Attention module}    & \checkmark               & -               & -               & \checkmark               & \checkmark               & -               & \checkmark               & -               & -               & \checkmark                & -                & \checkmark                \\
		\textbf{Pooling Method}      & Blur            & Blur            & Average         & Average         & Blur            & Average         & Average         & Blur            & Max             & Max              & Max              & Max              \\
		\textbf{Activation Function} & Mish            & Mish            & Mish            & Mish            & ReLU            & ReLU            & ReLU            & ReLU            & Mish            & Mish             & ReLU             & ReLU  \\          
		\hline
	\end{tabular}
\end{table*}

\section{Evaluation}\label{Evaluation}
We evaluate our method's 2D pose estimation in a number of contemporary datasets and with respect to state-of-the-art methods. We show that our exceptionally lightweight and straightforward technique outperforms other notably larger and complex deep learning architectures, which are computationally expensive. Our experiments were performed on five different datasets, the characteristics of which are presented below.

\subsection{Datasets}
\textbf{PANOPTIC} \cite{Simon_2017_CVPR} is an accurate large-scale human posture dataset with many instances of occluded subjects. We based our training set on three dataset sessions, \textit{office1, cello3} and \textit{tools1}. In accordance with the literature \cite{boukhayma20193d}, the training set of MPII+NZSL was also included \cite{Simon_2017_CVPR}  resulting into a total of 165000 training images. The evaluation was made on the testing set of MPII+NZSL. 

The \textbf{HO-3D} \cite{hampali2019ho} is a newly released markerless dataset, consisting of 10505 images in the training set. We augmented the dataset's images by flipping and rotating them by 0-90-180 degrees.

The \textbf{FreiHAND} \cite{zimmermann2019freihand} provides a multi-view hands dataset, recorded in front of a green screen and augmented with artificial background, resulting into a total of 130240 image instances.

The \textbf{LSMV} \cite{gomez2019large} provides images of hands from multiple points of view. The total of frames is 80000. 

The \textbf{SHP} \cite{zhang20163d} provides 3D pose annotations of a person's hand, performing various gestures in 18000 frames.

\begin{figure*}[!h]
	\centering
	\begin{adjustbox}{minipage=\linewidth,scale=1.02}
		\begin{subfigure}[b]{0.24\textwidth}
			\centering
			\includegraphics[width=\textwidth]{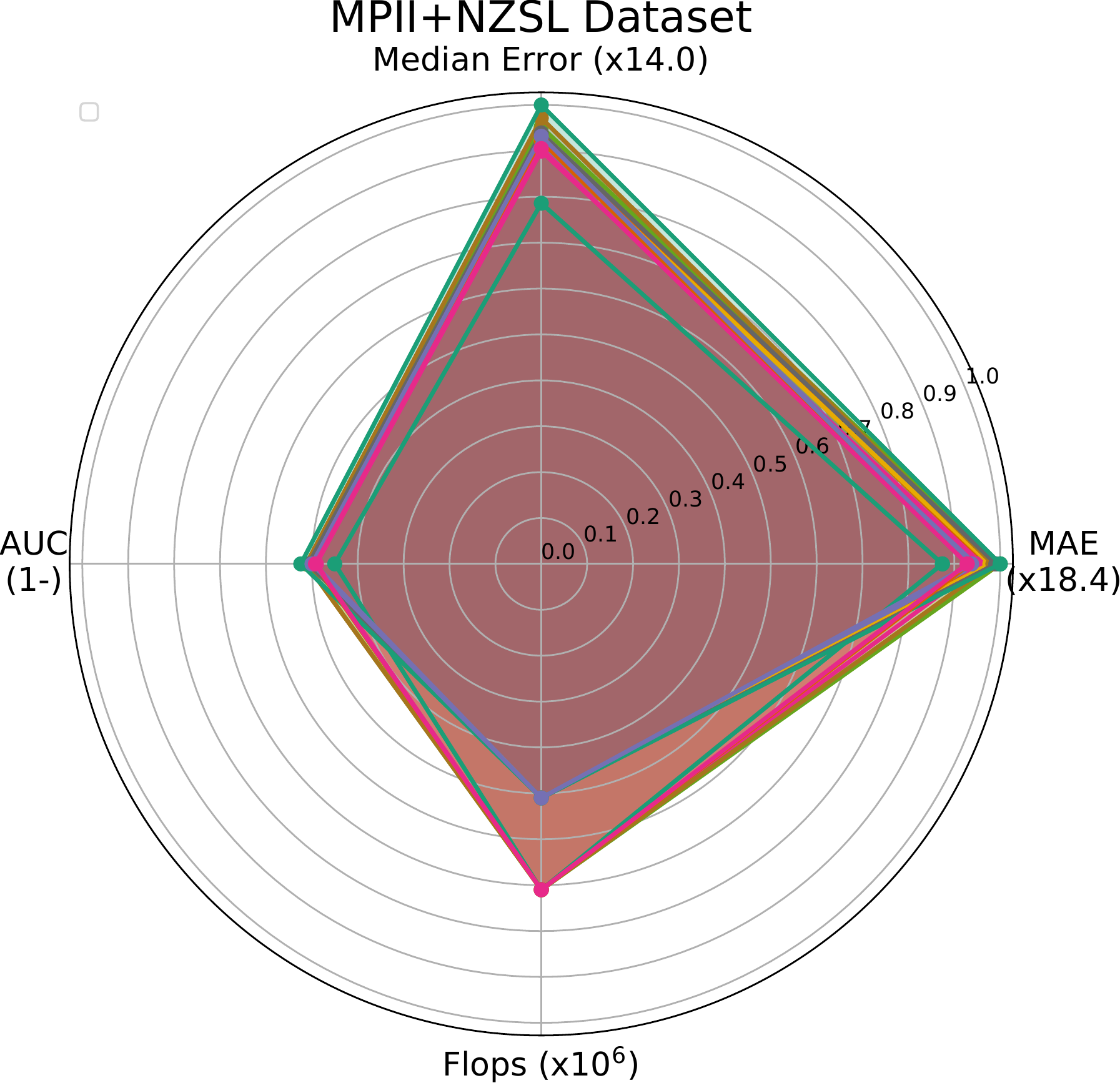}
			
		\end{subfigure}
		\begin{subfigure}[b]{0.24\textwidth}  
			\centering 
			\includegraphics[width=\textwidth]{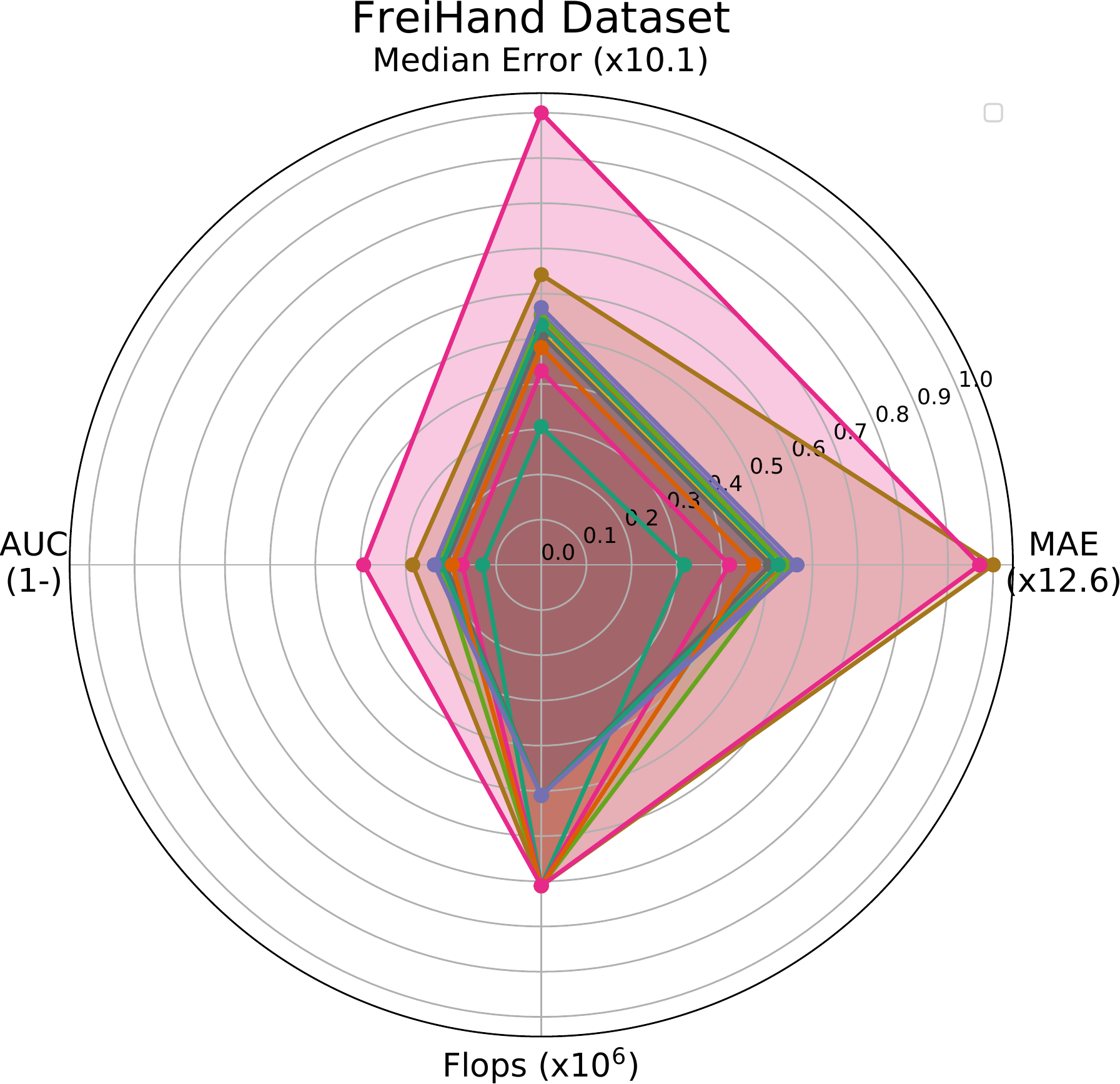}
			
		\end{subfigure}
		\begin{subfigure}[b]{0.24\textwidth}   
			\centering 
			\includegraphics[width=\textwidth]{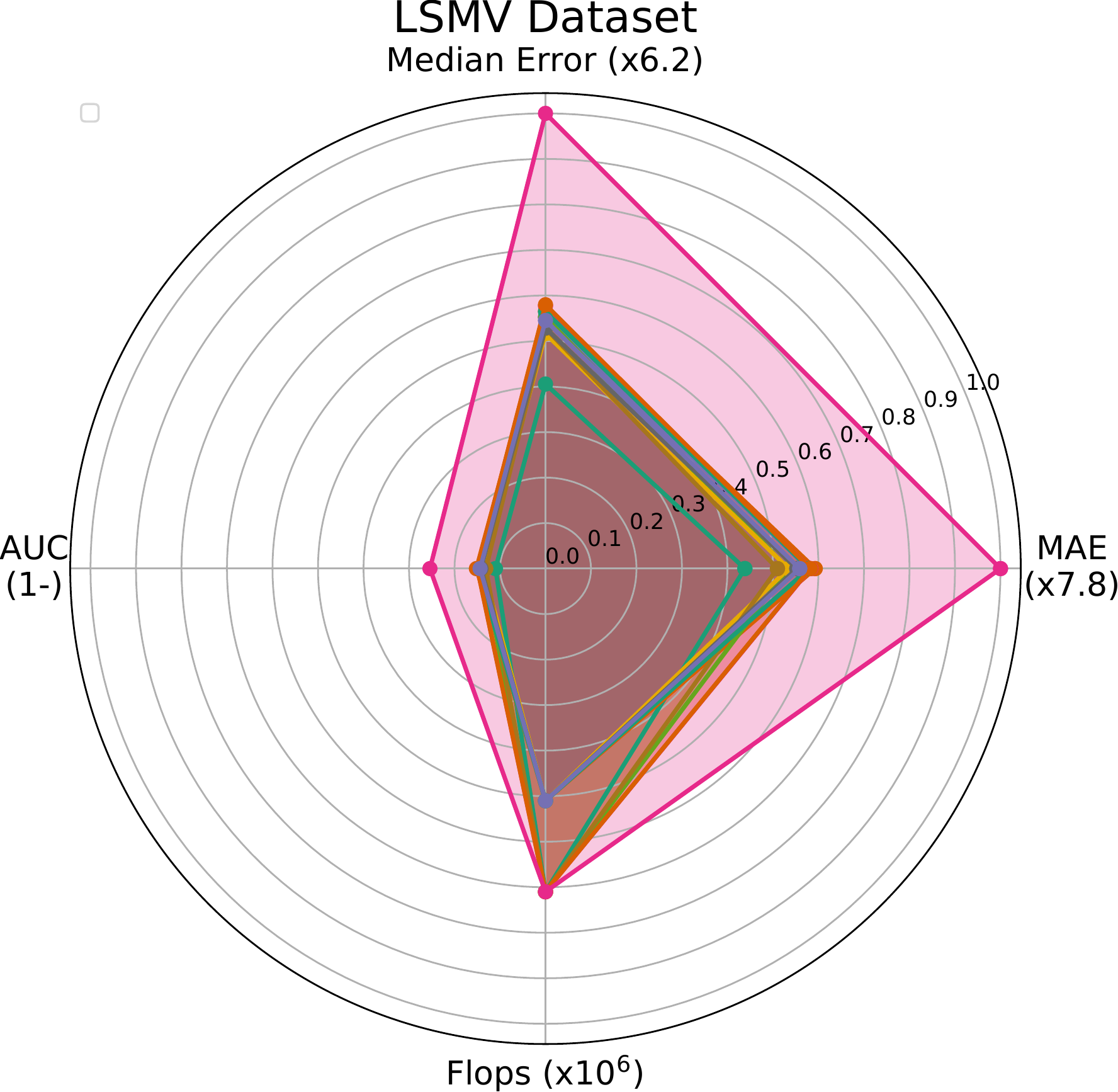}
			
		\end{subfigure}
		\begin{subfigure}[b]{0.24\textwidth}   
			\centering 
			\includegraphics[width=\textwidth]{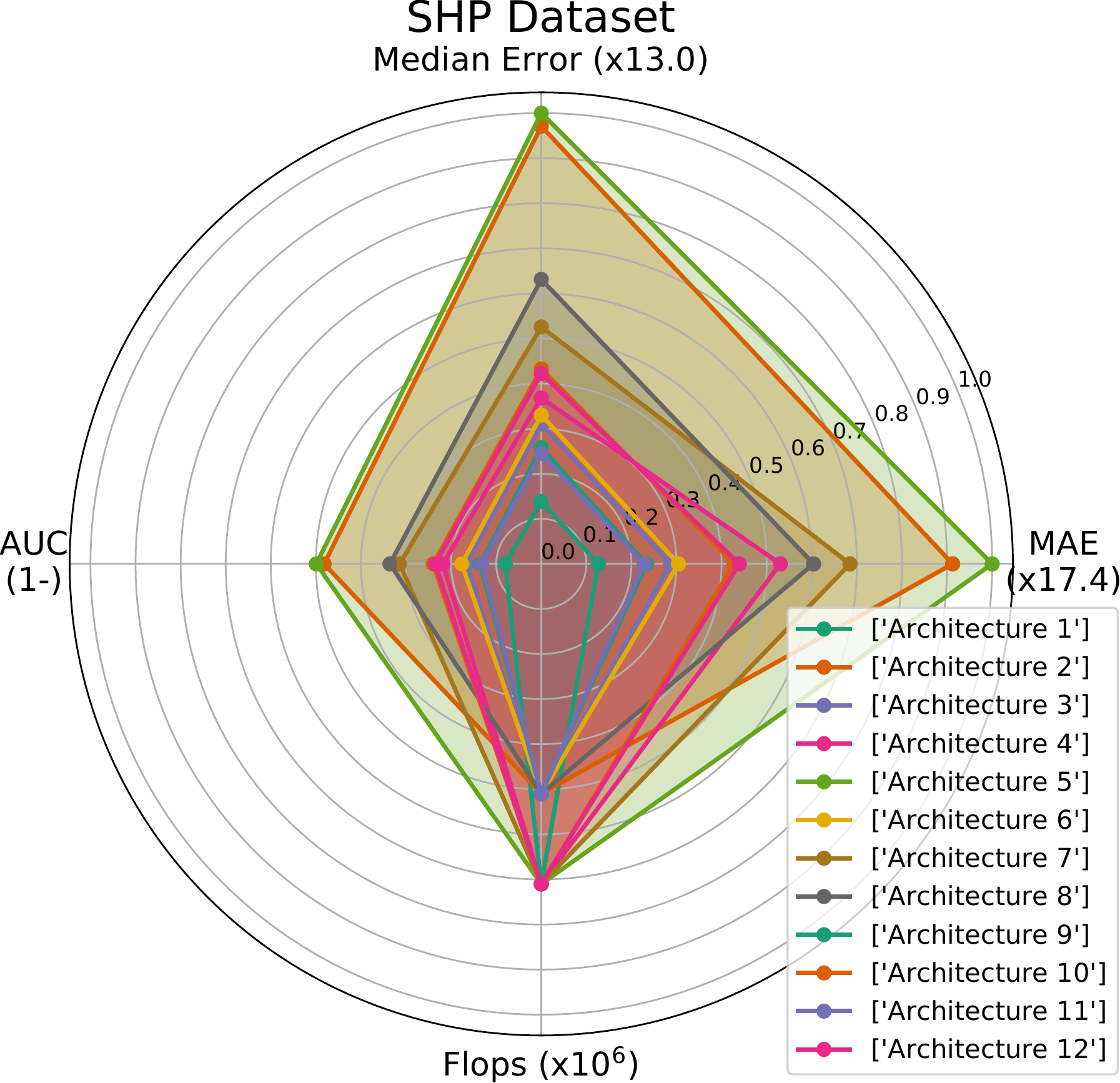}
			
		\end{subfigure}
	\caption{Results under different architecture configurations.}
		
		\label{fig:diff_radars}
	\end{adjustbox}
\end{figure*}

\begin{figure*}[!h]
	\centering
	\begin{adjustbox}{minipage=\linewidth,scale=1.02}
		\begin{subfigure}[b]{0.24\textwidth}
			\centering
			\includegraphics[width=\textwidth]{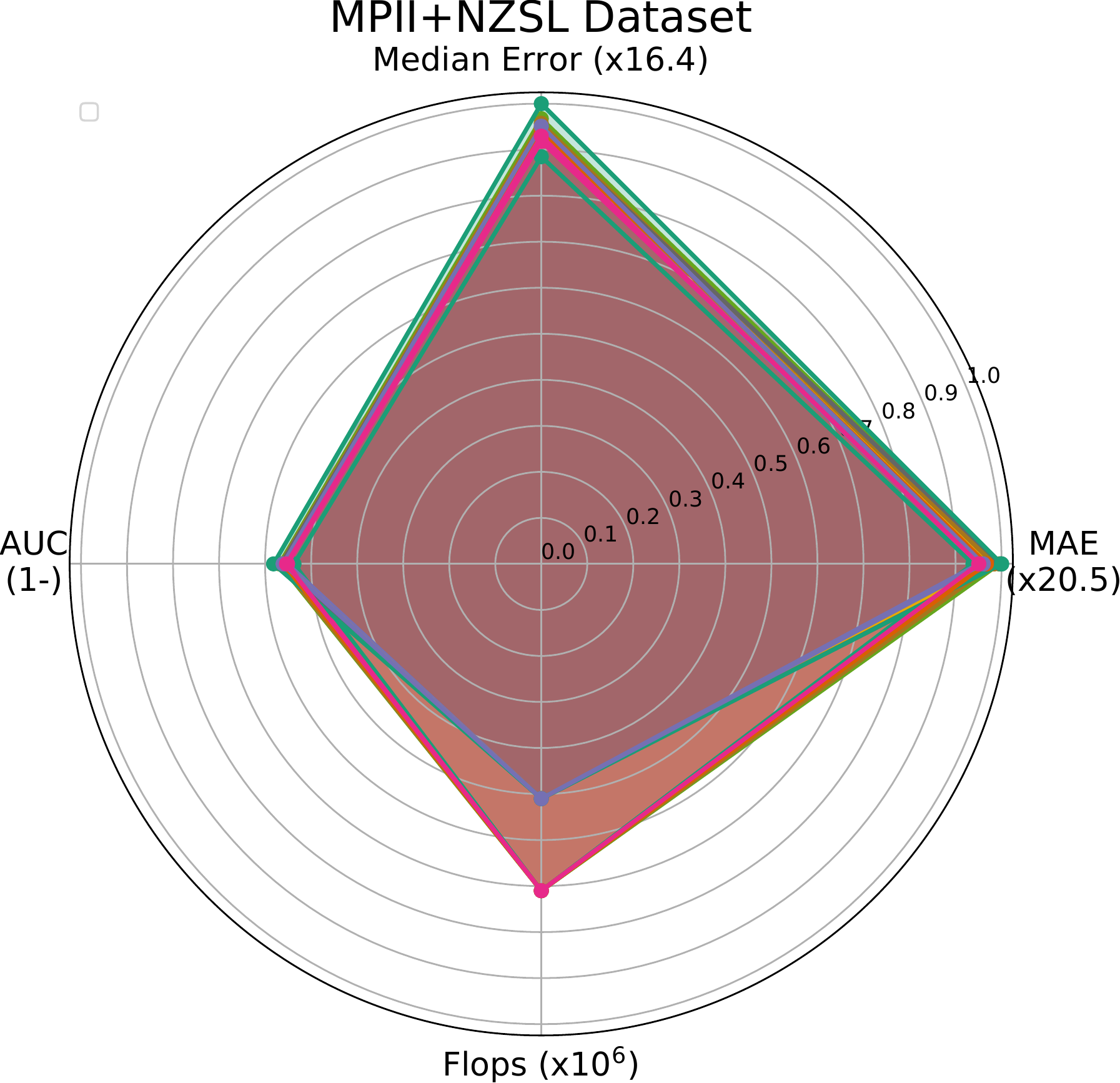}
			
		\end{subfigure}
		\begin{subfigure}[b]{0.24\textwidth}  
			\centering 
			\includegraphics[width=\textwidth]{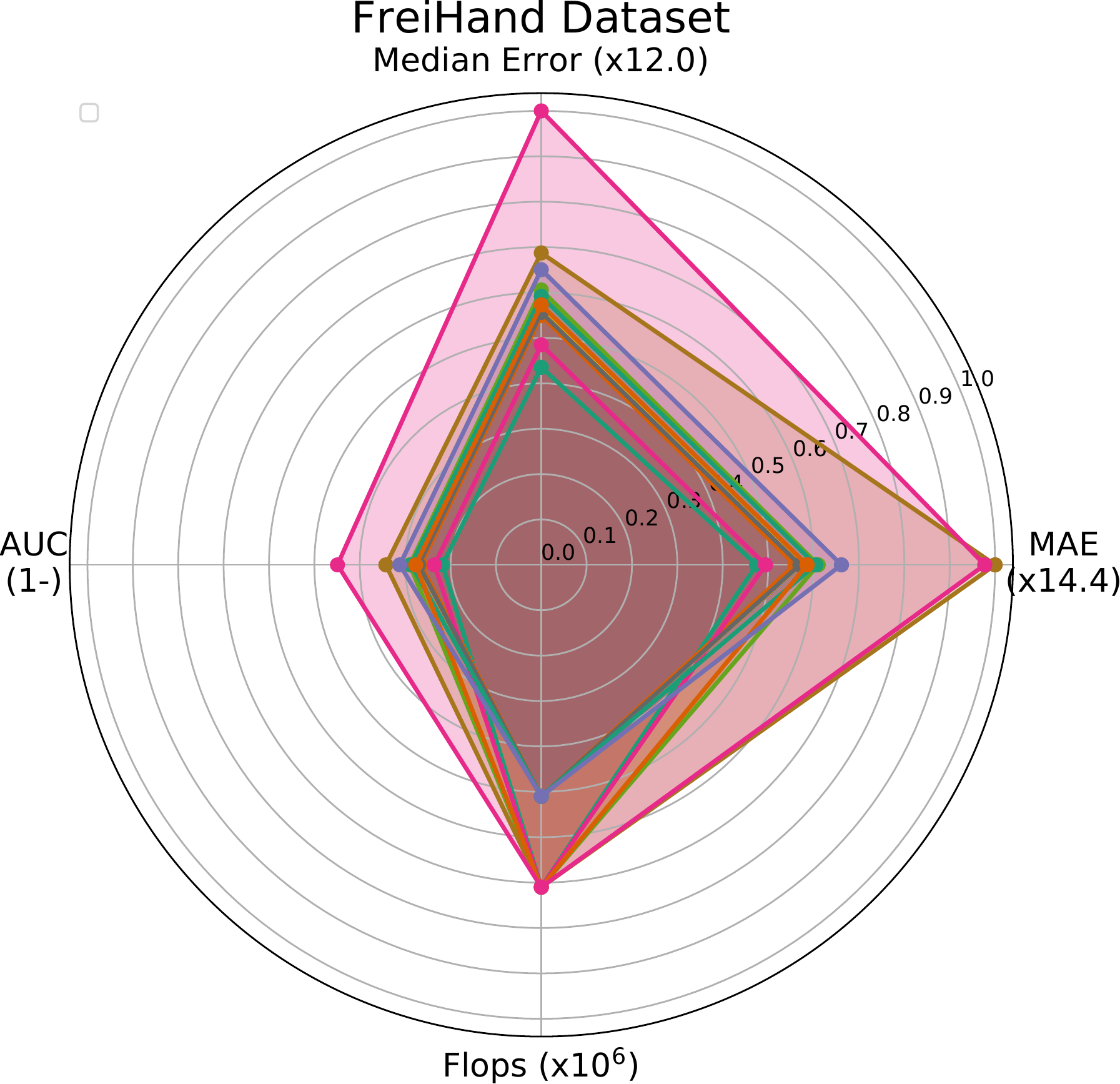}
			
		\end{subfigure}
		\begin{subfigure}[b]{0.24\textwidth}   
			\centering 
			\includegraphics[width=\textwidth]{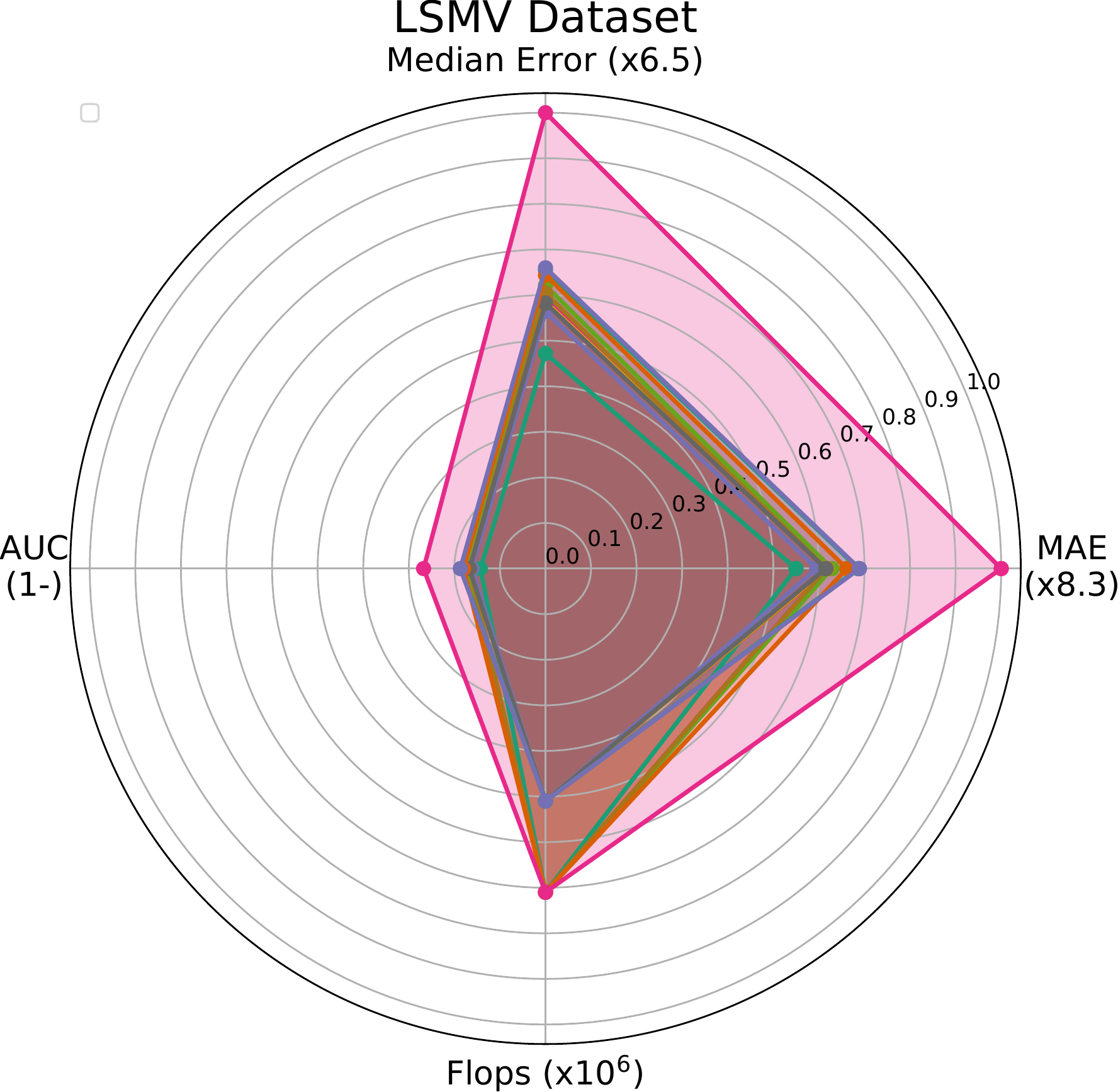}
			
		\end{subfigure}
		\begin{subfigure}[b]{0.24\textwidth}   
			\centering 
			\includegraphics[width=\textwidth]{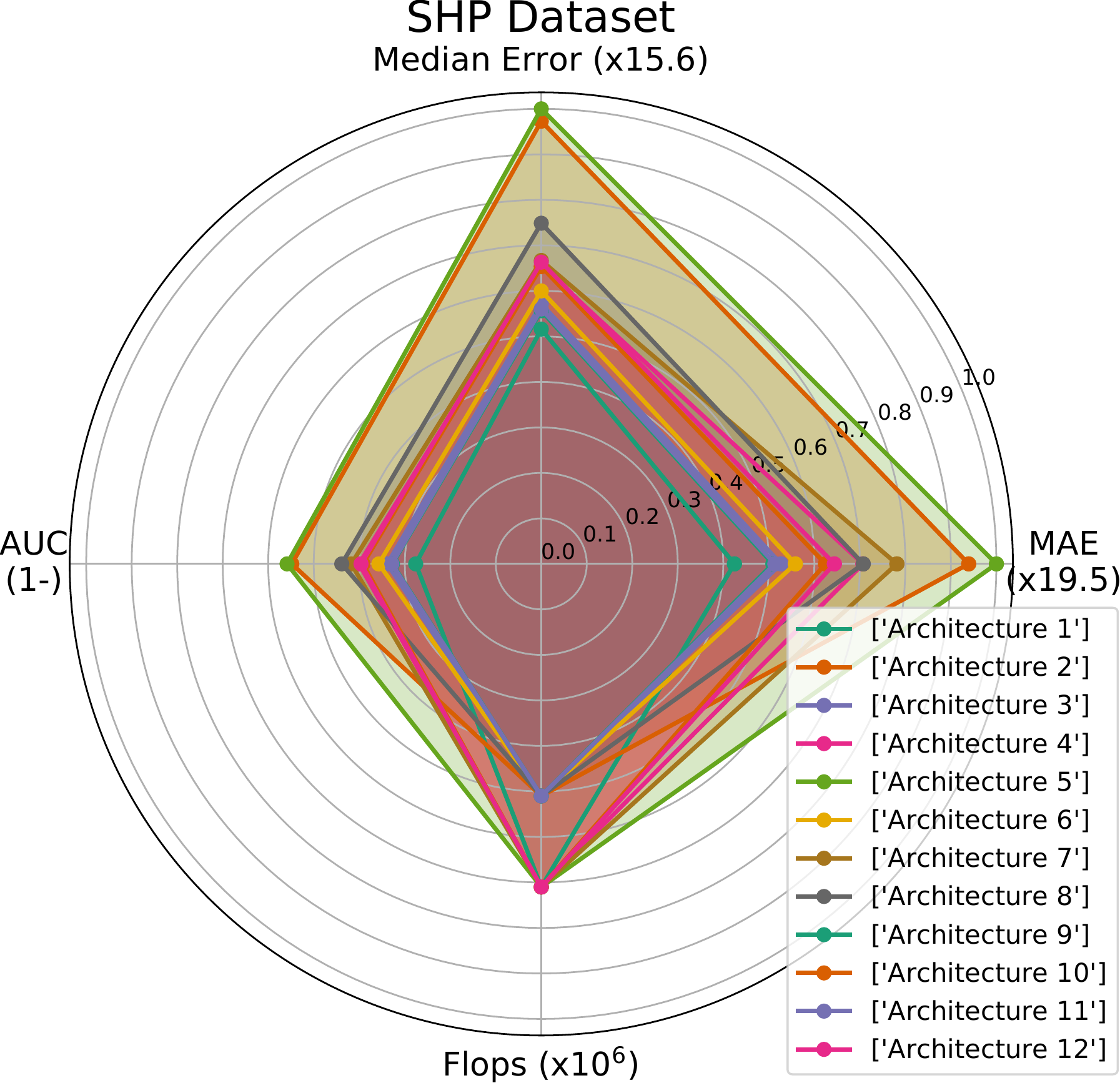}
			
		\end{subfigure}
		\caption{Results under different architecture configurations with randomly shifted input.}
		
		\label{fig:diff_radars_t}
	\end{adjustbox}
\end{figure*}
Each dataset was separately evaluated and split by a rule of 80\%-10\%-10\% for training, validation and testing, respectively. Every image was cropped to the resolution of $224\times 224$.
\subsection{Ablation studies}
In this subsection, we wish to justify the architectural blocks identified into our system. We evaluate the proposed method under different settings by:  i) optionally excluding the attention augmented convolution module, ii) using different pooling methods and iii) trying different activation functions, as presented in Table {{\ref{table:diff_archs}}}. Regarding the computational efficiency of our approaches, adding the attention augmented convolution module leads to a slight increment of the overall FLOPs to \textit{7.1 Million}. Each of these twelve configurations were trained and tested on the same four datasets, used for evaluation, including {PANOPTIC} \cite{Simon_2017_CVPR}, {FreiHAND} \cite{zimmermann2019freihand}, {LSMV} \cite{gomez2019large} and {SHP} \cite{zhang20163d}. 

The results for each combination are presented in \figurename{ \ref{fig:diff_radars}}, where the supremacy of \textit{Architecture 1} over every other one becomes apparent, as it is the one occupying the minimum area in all radar-charts. In order to prove the robustness of the proposed method to input's translations, we also applied random shifts to each of the corresponding datasets during evaluation only, in both vertical and horizontal axis, with an interval of 20 pixels. As one can observe based on the results presented in \figurename{ \ref{fig:diff_radars_t}}, the input's translation leads into accuracy degradation for every tested architecture. However, \textit{Architecture 1} exhibits the modest decrease justifying its employment into our final architecture. We attribute this behavior to the \textit{Blur Pooling layer}, which acts as an anti-aliasing filter during sub-sampling. This allows the network to propagate as much information as possible to the deeper layers improving the regression results. It is also worth mentioning that, instead of the typical \textit{ReLU} non-linearity, the combination of \textit{Blur Pooling} with the \textit{Mish} activation function, yields the highest overall system performance.

\subsection{Comparative results}
\begin{figure*}[!t]
	\centering
	
	\begin{subfigure}[c]{\sze\linewidth}
		\includegraphics[width=\linewidth]{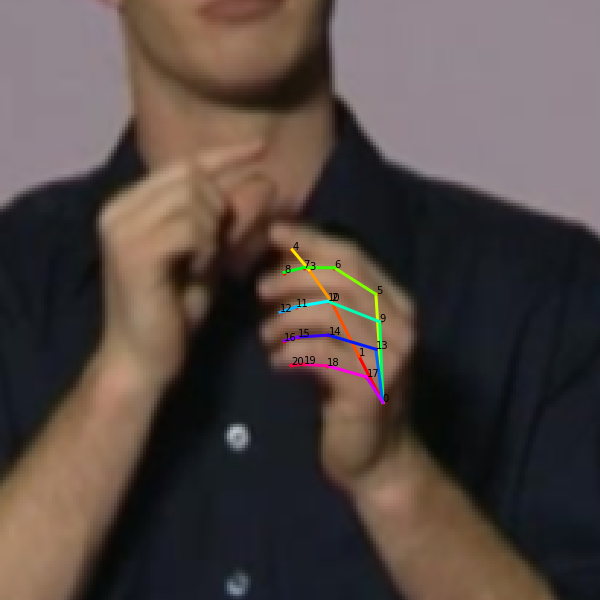}
	\end{subfigure}
	\hspace{\hor}
	\begin{subfigure}[c]{\sze\linewidth}
		\includegraphics[width=\linewidth]{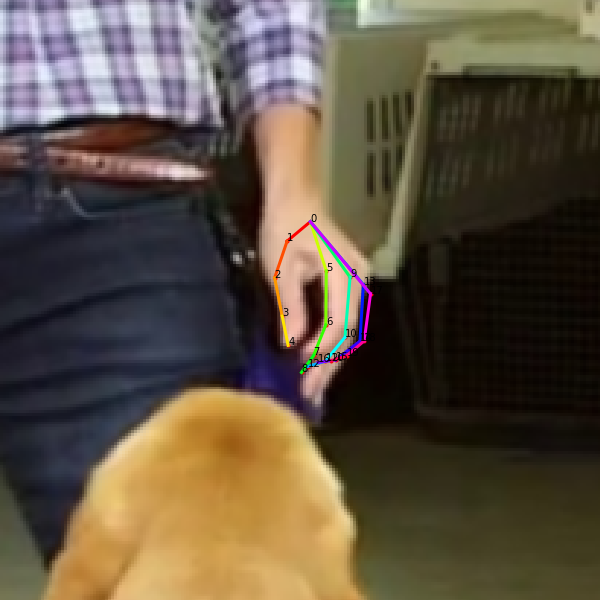}
	\end{subfigure}
	\hspace{\hor}
	\begin{subfigure}[c]{\sze\linewidth}
		\includegraphics[width=\linewidth]{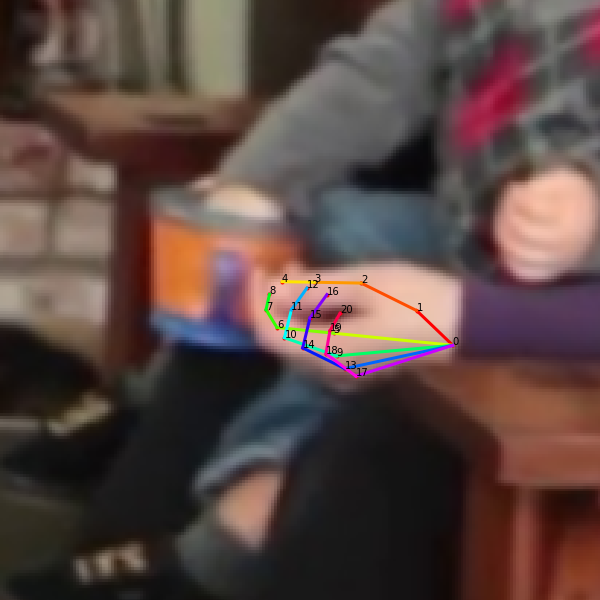}
	\end{subfigure}
	\hspace{\hor}
	\begin{subfigure}[c]{\sze\linewidth}
		\includegraphics[width=\linewidth]{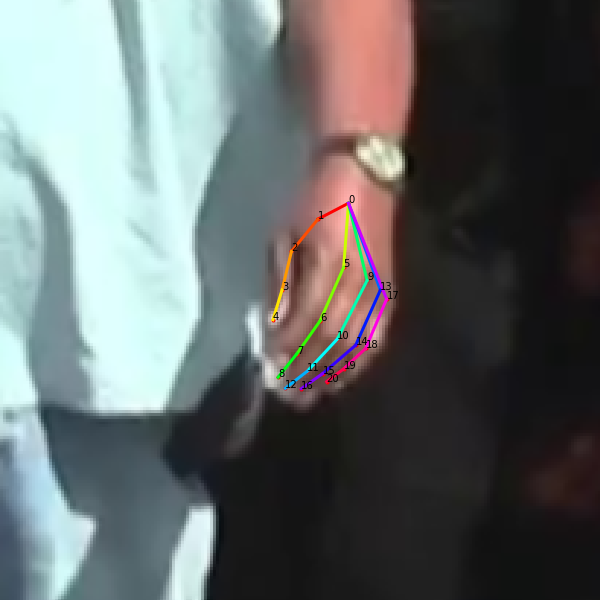}
	\end{subfigure}
	\hspace{\hor}
	\begin{subfigure}[c]{\sze\linewidth}
		\includegraphics[width=\linewidth]{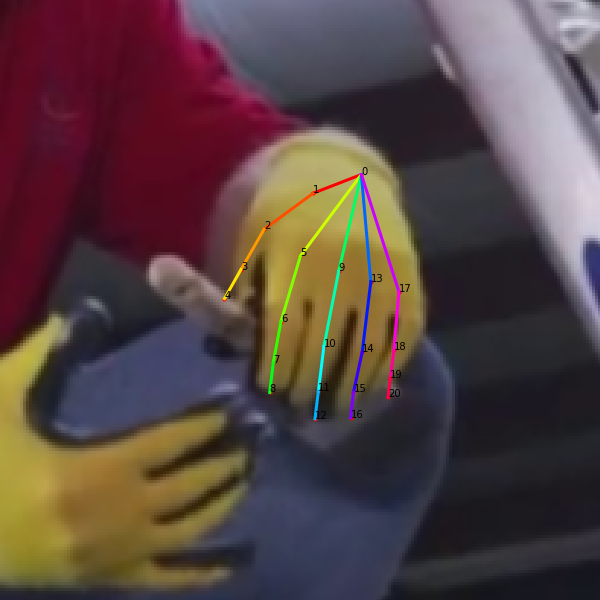}
	\end{subfigure}
	\hspace{\hor}
	\begin{subfigure}[c]{\sze\linewidth}
		\includegraphics[width=\linewidth]{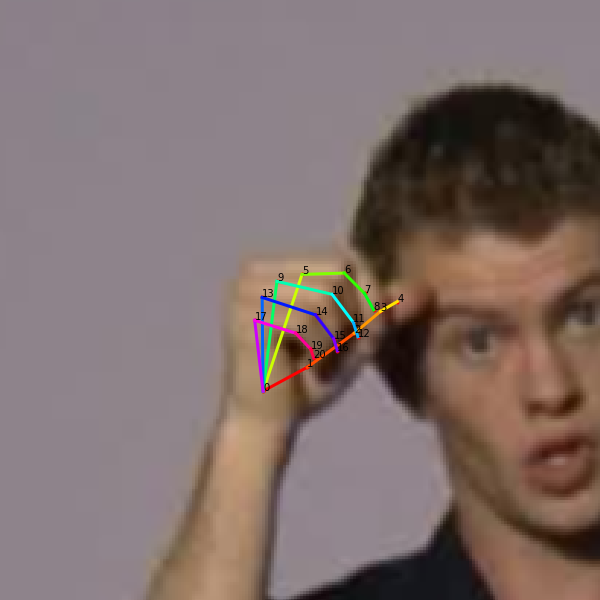}
	\end{subfigure}

	\vspace{\ver}	
	\begin{subfigure}[c]{\sze\linewidth}
		\includegraphics[width=\linewidth]{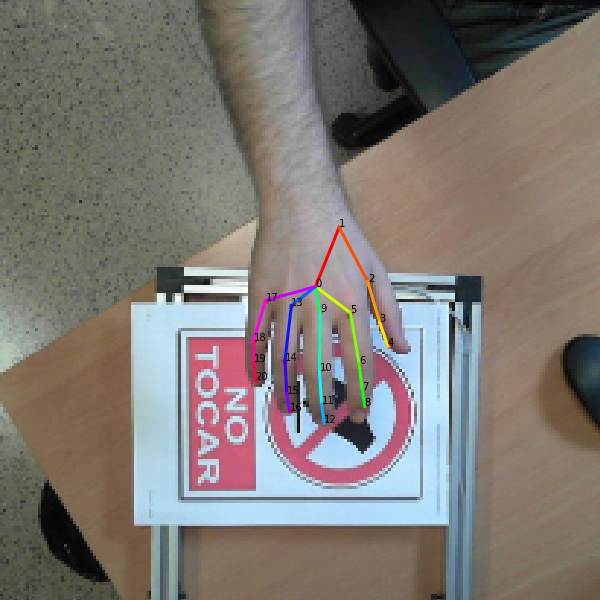}
	\end{subfigure}
	\hspace{\hor}
	\begin{subfigure}[c]{\sze\linewidth}
		\includegraphics[width=\linewidth]{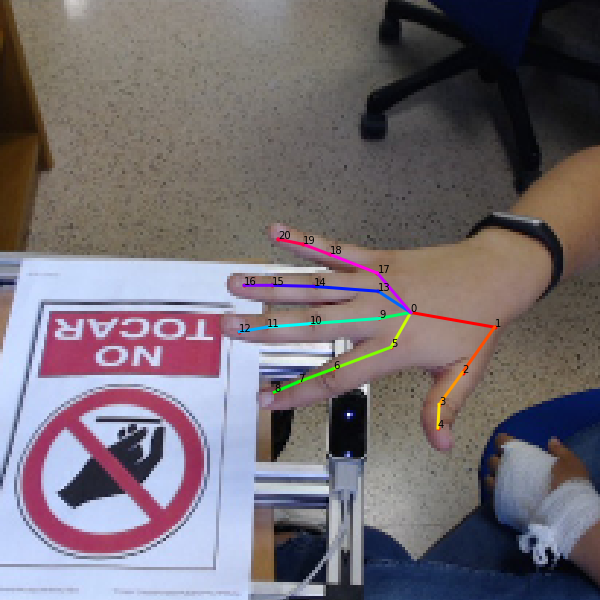}
	\end{subfigure}
	\hspace{\hor}
	\begin{subfigure}[c]{\sze\linewidth}
		\includegraphics[width=\linewidth]{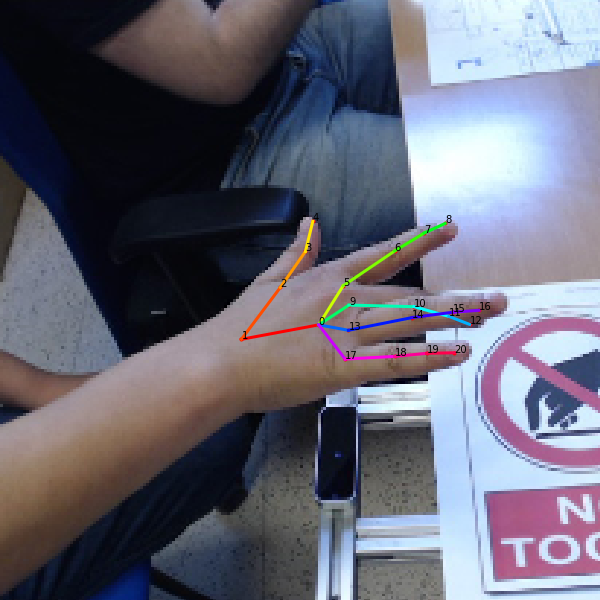}
	\end{subfigure}
	\hspace{\hor}
	\begin{subfigure}[c]{\sze\linewidth}
		\includegraphics[width=\linewidth]{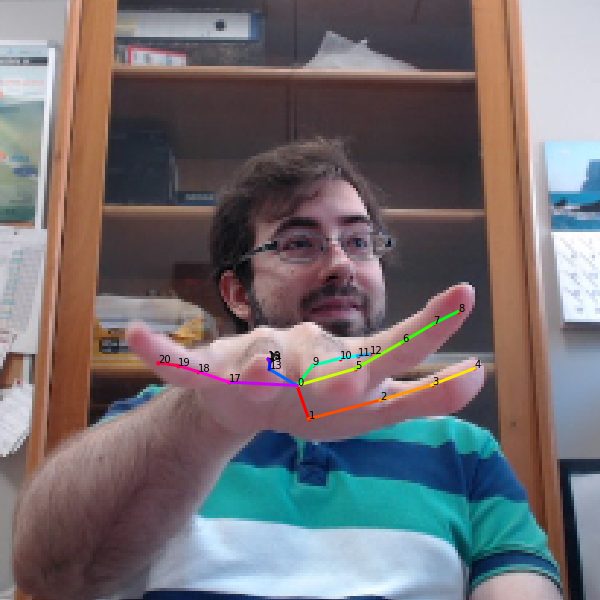}
	\end{subfigure}
	\hspace{\hor}
	\begin{subfigure}[c]{\sze\linewidth}
		\includegraphics[width=\linewidth]{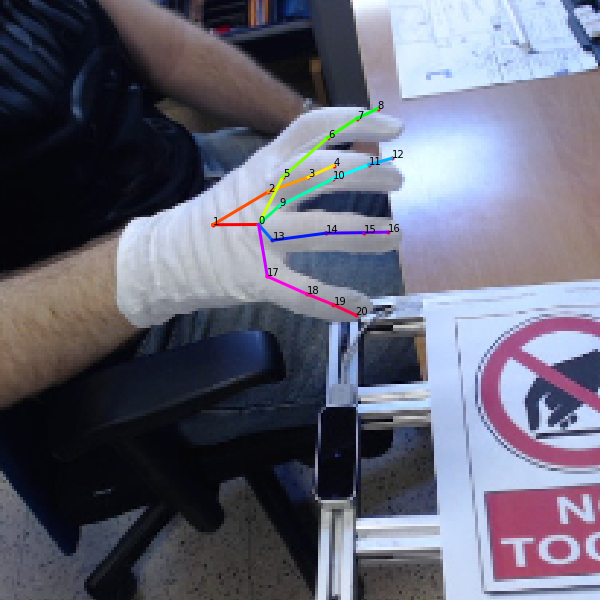}
	\end{subfigure}
	\hspace{\hor}
	\begin{subfigure}[c]{\sze\linewidth}
		\includegraphics[width=\linewidth]{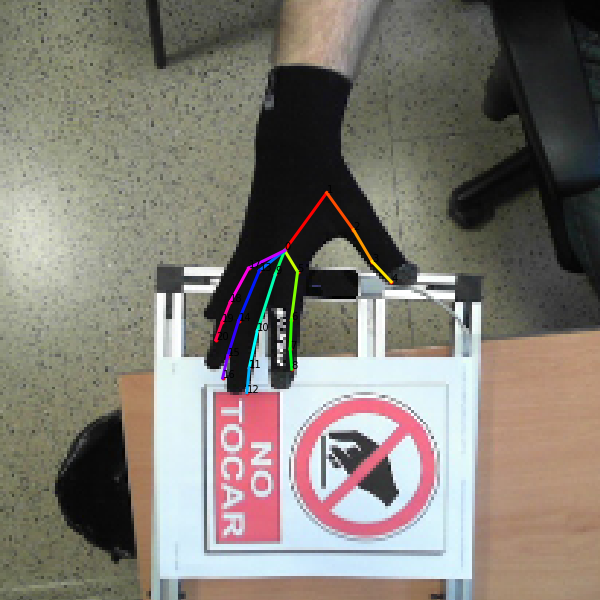}
	\end{subfigure}

	\vspace{\ver}	
	\begin{subfigure}[c]{\sze\linewidth}
		\includegraphics[width=\linewidth]{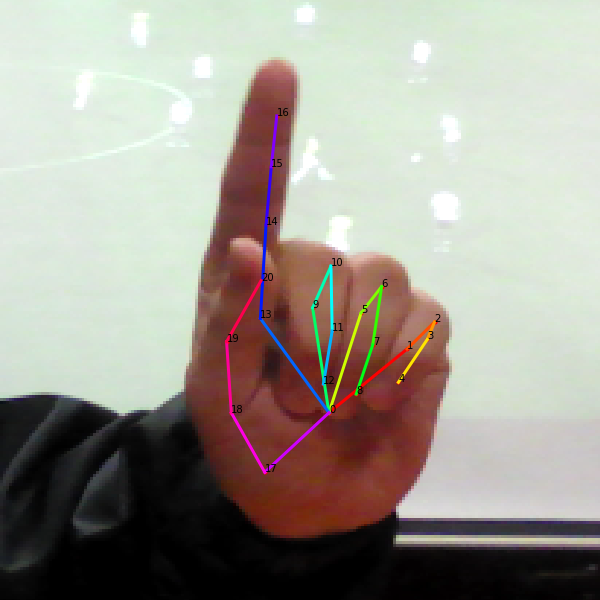}
	\end{subfigure}
	\hspace{\hor}
	\begin{subfigure}[c]{\sze\linewidth}
		\includegraphics[width=\linewidth]{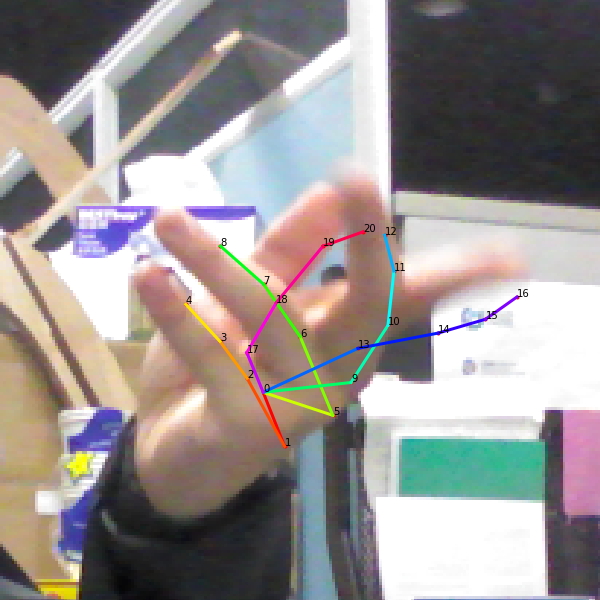}
	\end{subfigure}
	\hspace{\hor}
	\begin{subfigure}[c]{\sze\linewidth}
		\includegraphics[width=\linewidth]{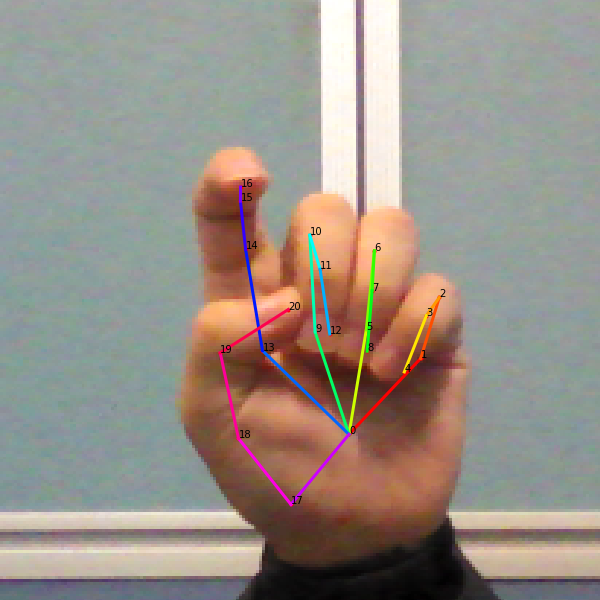}
	\end{subfigure}
	\hspace{\hor}
	\begin{subfigure}[c]{\sze\linewidth}
		\includegraphics[width=\linewidth]{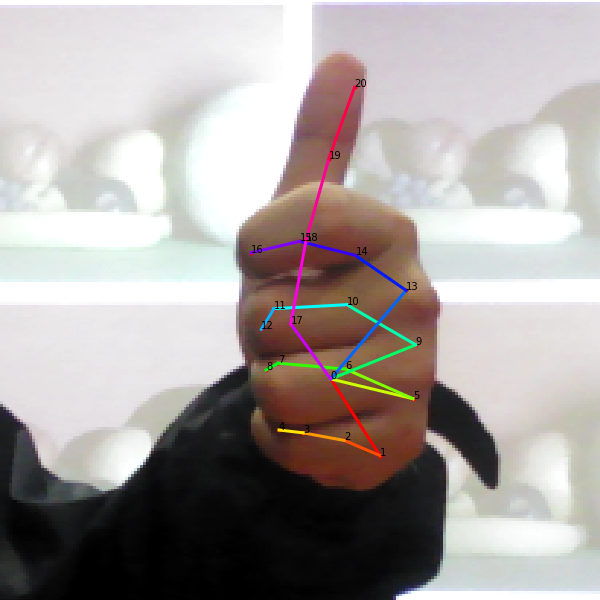}
	\end{subfigure}
	\hspace{\hor}
	\begin{subfigure}[c]{\sze\linewidth}
		\includegraphics[width=\linewidth]{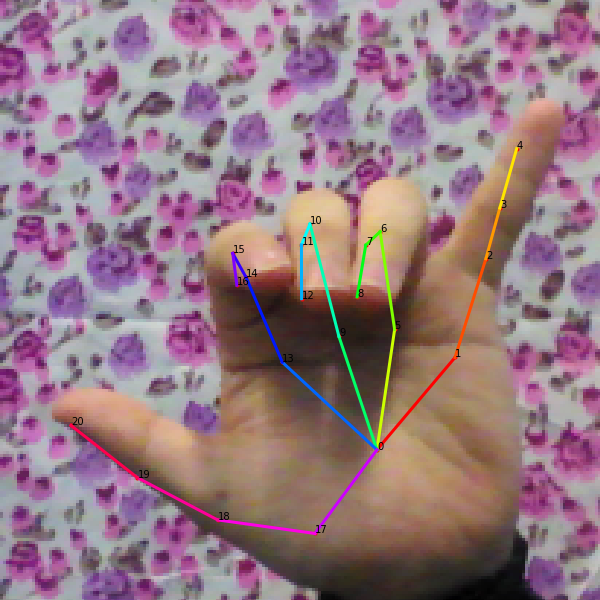}
	\end{subfigure}
	\hspace{\hor}
	\begin{subfigure}[c]{\sze\linewidth}
		\includegraphics[width=\linewidth]{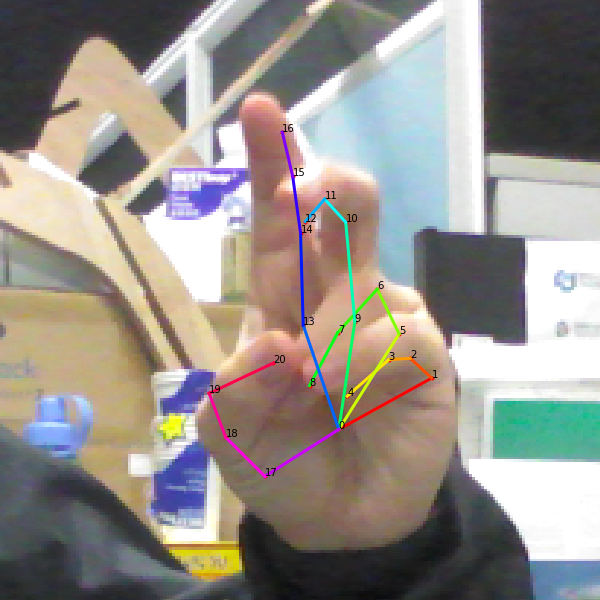}
	\end{subfigure}

	\vspace{\ver}	
	\begin{subfigure}[c]{\sze\linewidth}
		\includegraphics[width=\linewidth]{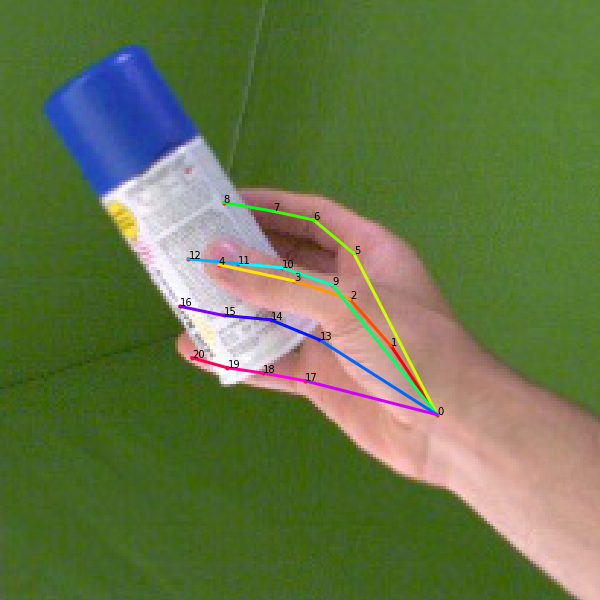}
	\end{subfigure}
	\hspace{\hor}
	\begin{subfigure}[c]{\sze\linewidth}
		\includegraphics[width=\linewidth]{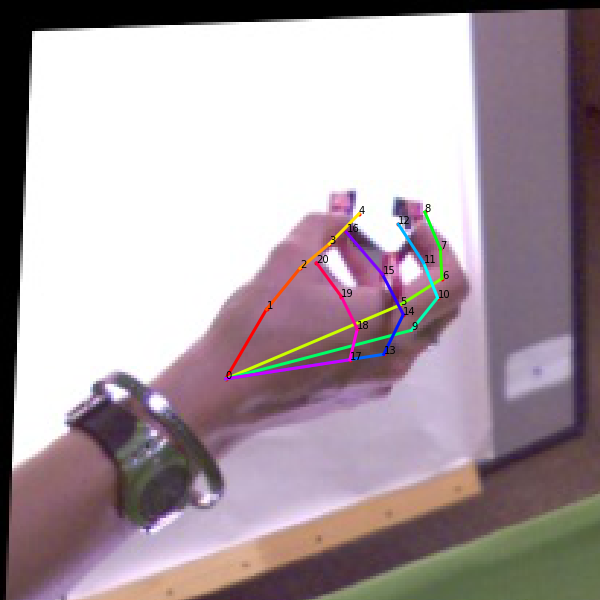}
	\end{subfigure}
	\hspace{\hor}
	\begin{subfigure}[c]{\sze\linewidth}
		\includegraphics[width=\linewidth]{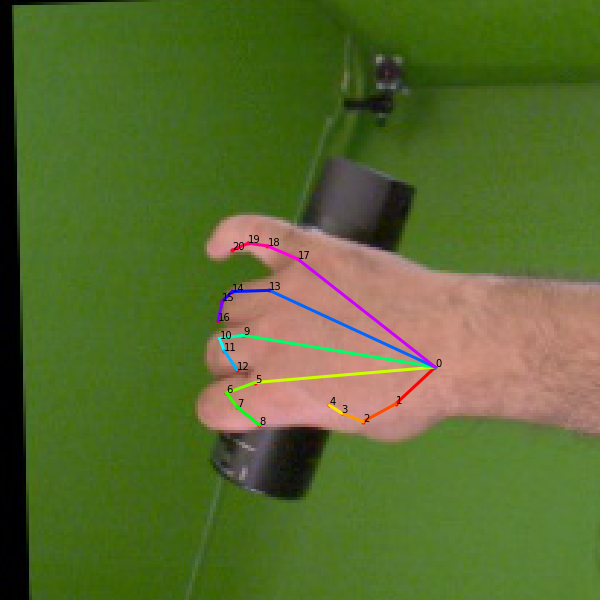}
	\end{subfigure}
	\hspace{\hor}
	\begin{subfigure}[c]{\sze\linewidth}
		\includegraphics[width=\linewidth]{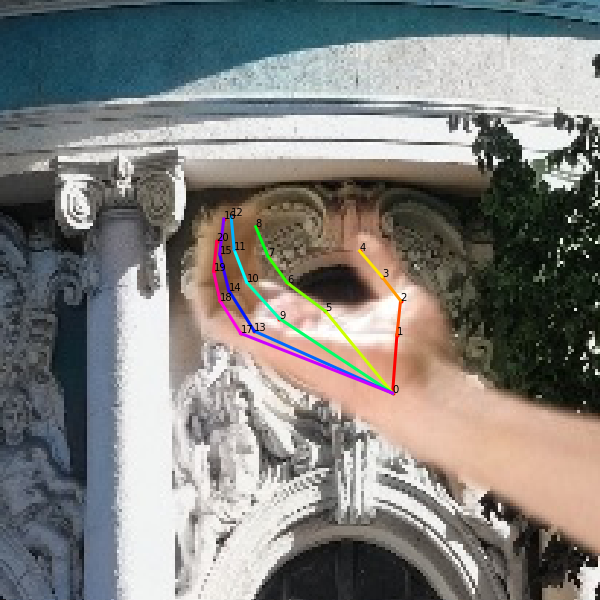}
	\end{subfigure}
	\hspace{\hor}
	\begin{subfigure}[c]{\sze\linewidth}
		\includegraphics[width=\linewidth]{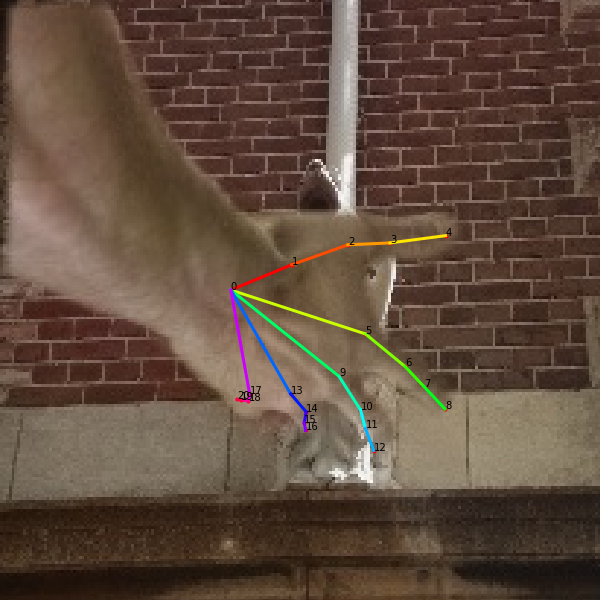}
	\end{subfigure}
	\hspace{\hor}
	\begin{subfigure}[c]{\sze\linewidth}
		\includegraphics[width=\linewidth]{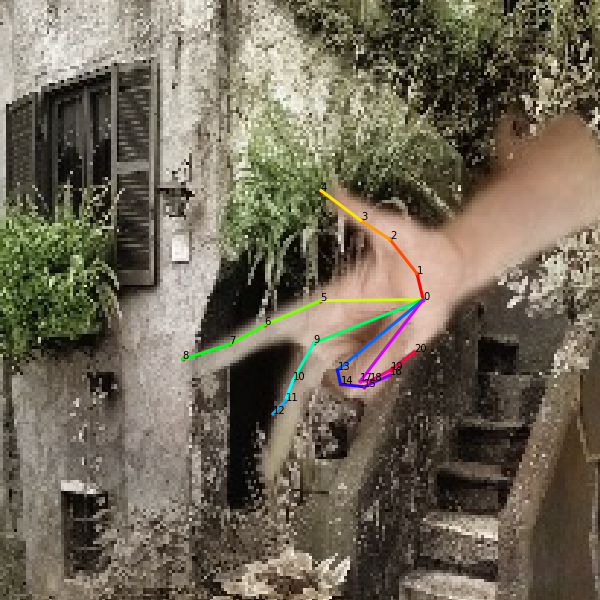}
	\end{subfigure}

	\vspace{\ver}	
	\begin{subfigure}[c]{\sze\linewidth}
		\includegraphics[width=\linewidth]{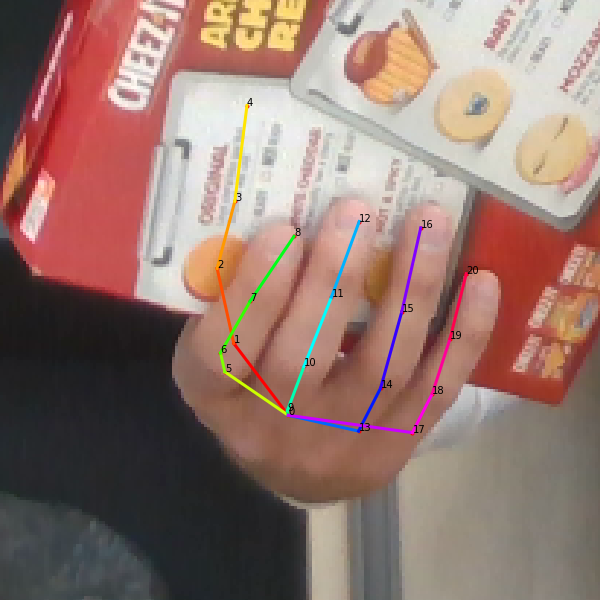}
	\end{subfigure}
	\hspace{\hor}
	\begin{subfigure}[c]{\sze\linewidth}
		\includegraphics[width=\linewidth]{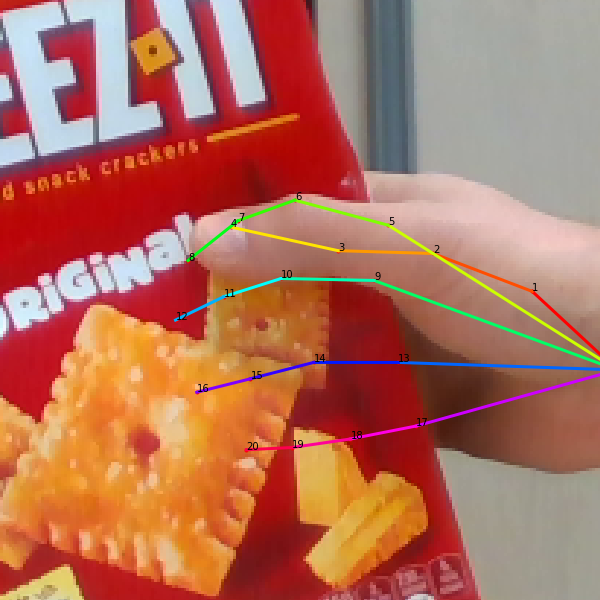}
	\end{subfigure}
	\hspace{\hor}
	\begin{subfigure}[c]{\sze\linewidth}
		\includegraphics[width=\linewidth]{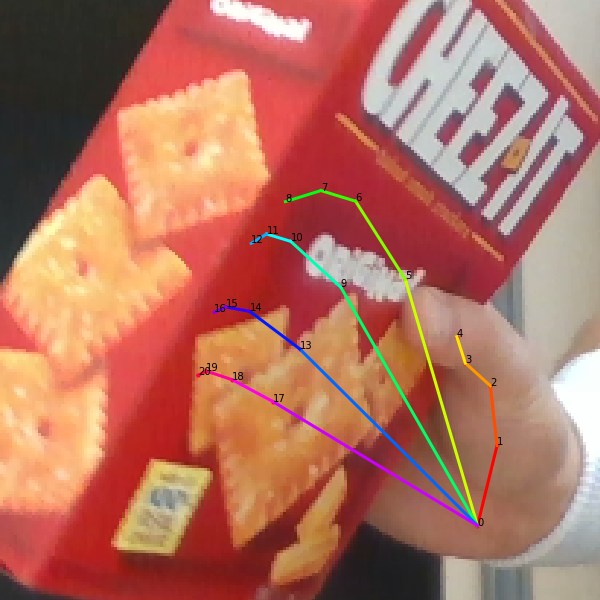}
	\end{subfigure}
	\hspace{\hor}
	\begin{subfigure}[c]{\sze\linewidth}
		\includegraphics[width=\linewidth]{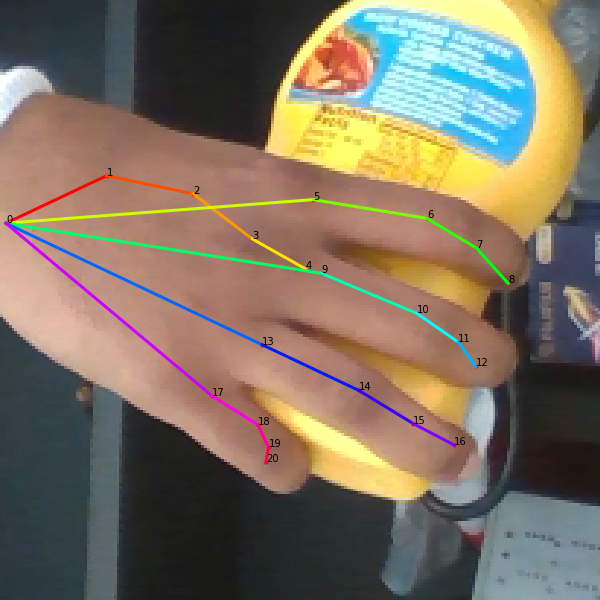}
	\end{subfigure}
	\hspace{\hor}
	\begin{subfigure}[c]{\sze\linewidth}
		\includegraphics[width=\linewidth]{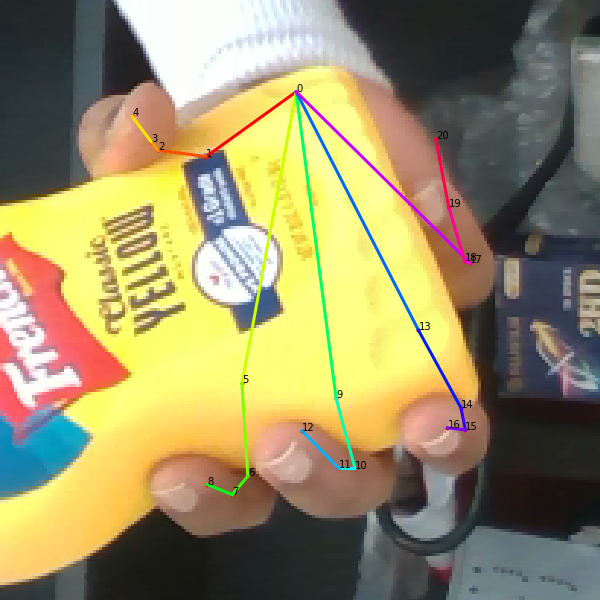}
	\end{subfigure}
	\hspace{\hor}
	\begin{subfigure}[c]{\sze\linewidth}
		\includegraphics[width=\linewidth]{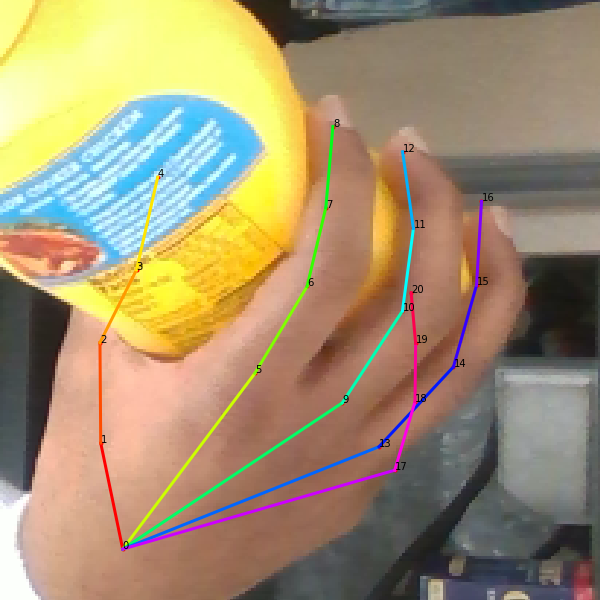}
	\end{subfigure}
	
	\caption{Our 2D hand pose estimations on different testing sets.}
	\label{fig:photo}
\end{figure*}

We compare our results with other state-of-the-art methods in \figurename{ \ref{evaluation}}, according to the protocol proposed in \cite{Simon_2017_CVPR}, showing that our method outperforms other approaches. More specifically, in \figurename{ \ref{grapha}} and \figurename{ \ref{graphb}}, the percentage of correct keypoints is visualized for different absolute and normalized thresholds, respectively, and compared to other techniques. Figure {\ref{graphc}} depicts our method's performance trained on different datasets. The abovementioned results are also summarized in Table \ref{table:results}. 
\figurename{ \ref{heatmaps}}, contains representative examples of the feature maps, as computed by our proposed \textit{Attention Augmented Inverted Bottleneck Block} for a sample of images, during different network's stages. Even if the network was not explicitly trained for the generation of a heat map, it is easily perceptible the effectiveness of the specific block to detect the points of interest, which in our case are the hand's keypoints.  Finally, in  \figurename{ \ref{fig:photo}}, some qualitative results of our method are presented for different dataset instances. 

\section{Conclusions}\label{Conclusions}

We presented an alternative approach in contrast to the majority of pose estimation methods which incorporate complex and computationally inefficient architectures. Our proposed single-stage end-to-end CNN model exhibits competitive results with just \textit{1.9M} parameters and a model's size of 11 Mbytes, which is achieved by directly predicting the joints' coordinates. This property allows for the deployment of our system on low processing power devices accommodating the operational guidelines of modern mobile systems.

The method's success is mainly based on the effectiveness of the proposed \textit{Attention Augmented Inverted Bottleneck Block} to understand global constraints and correlations between keypoints, as well as, to the architecture's ability to share a ``collective knowledge`` among the subsequent layers. Closing, it is highly possible that our proposed architecture could be suitable for other tasks too, such as 3D pose estimation, human body pose estimation or classification, which we intend to explore as a future work.

\section{Acknowledgments}
This work was supported by Google's TensorFlow Research Cloud and Google's Research Credits programme.

\bibliographystyle{IEEEtran}
\bibliography{refs.bib}

\vskip -2\baselineskip plus -1fil
\begin{IEEEbiography}
[{\includegraphics[width=1in,height=1.25in,clip, keepaspectratio]{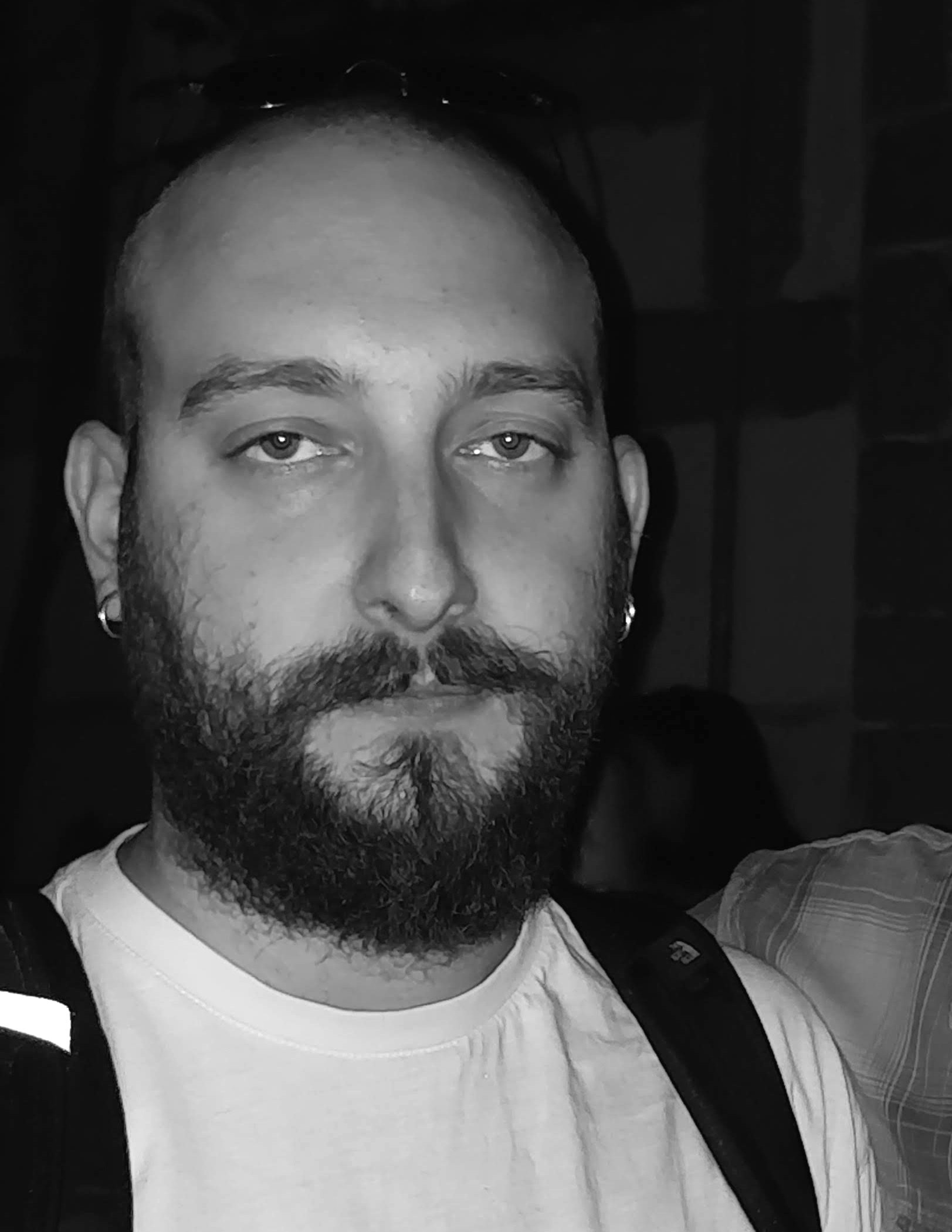}}]{Nicholas Santavas}
received the diploma degree from the Department of Production and Management Engineering, Democritus University of Thrace, Greece, in 2018. Currently, he is working toward the PhD degree in the Laboratory of Robotics and Automation (LRA), Department of Production and Management Engineering, Democritus University of Thrace (DUTH), Greece. His research interest include Artificial Neural Networks and Human-Computer Interaction applications
\end{IEEEbiography}
\vskip -3\baselineskip plus -1fil
\begin{IEEEbiography}
[{\includegraphics[width=1in,height=1.25in,clip, keepaspectratio]{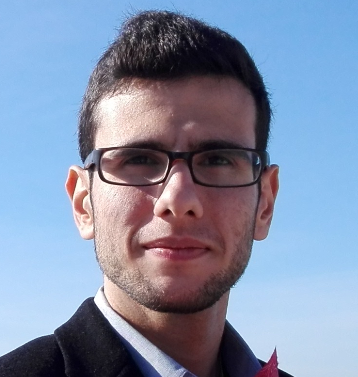}}]{Ioannis Kansizoglou}
received the diploma degree from the Department of Electrical and Computer Engineering, Aristotle University of Thessaloniki, Greece, in 2017. Currently, he is working toward the PhD degree in the Laboratory of Robotics and Automation (LRA), Department of Production and Management Engineering, Democritus University of Thrace (DUTH), Greece, working on emotion analysis and its application in robotics.
\end{IEEEbiography}
\vskip -3\baselineskip plus -1fil
\begin{IEEEbiography}
[{\includegraphics[width=1in,height=1.25in,clip, keepaspectratio]{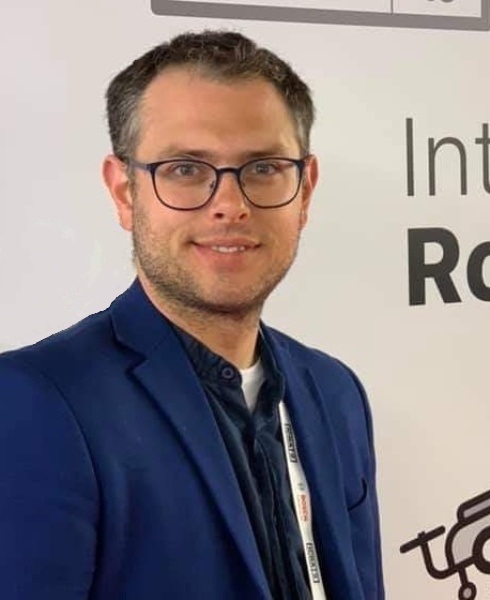}}]{Loukas Bampis}
received the diploma degree in electrical and computer engineering and PhD degree in machine vision and embedded systems from the Democritus University of Thrace (DUTH), Greece, in 2013 and 2019, respectively. He is currently a postdoctoral fellow in the Laboratory of Robotics and Automation (LRA), Department of Production and Management Engineering, DUTH. His work has been supported through several research projects funded by the European Space Agency, the European Commission and the Greek government. His research interests include real-time localization and place recognition techniques using hardware accelerators and parallel processing.
\end{IEEEbiography}
\vskip -2\baselineskip plus -1fil
\begin{IEEEbiography}
[{\includegraphics[width=1in,height=1.25in,clip, keepaspectratio]{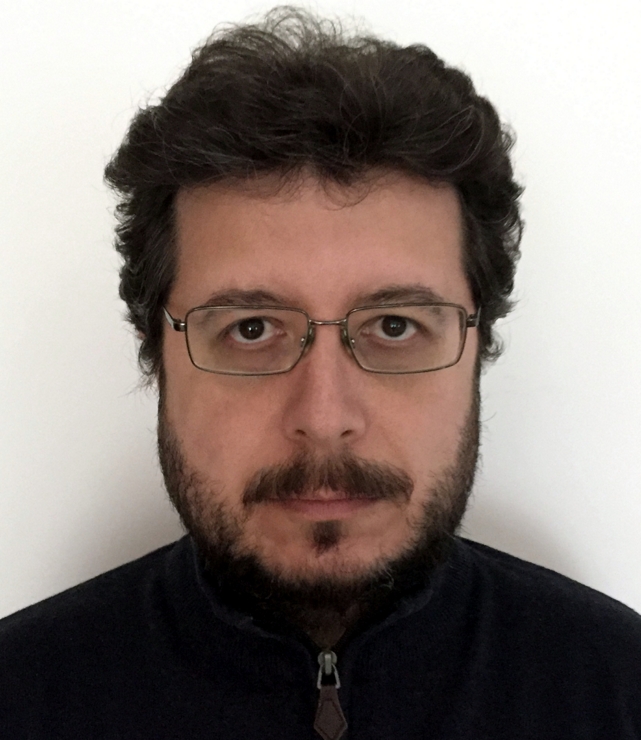}}]{Evangelos Karakasis}
received the Ph.D. degree in 2014 and the Diploma degree in production and management engineering in 2005 from the Democritus University of Thrace (DUTH), Xanthi, Greece.  His current research interests include computer vision, signal and image processing, pattern recognition and classification, machine learning/deep learning, anomaly detection, special functions and predictive maintenance. He has co-authored more than 30 papers in international journals, conferences and book chapters.
\end{IEEEbiography}
\vskip -2\baselineskip plus -1fil
\begin{IEEEbiography}
[{\includegraphics[width=1in,height=1.25in,clip, keepaspectratio]{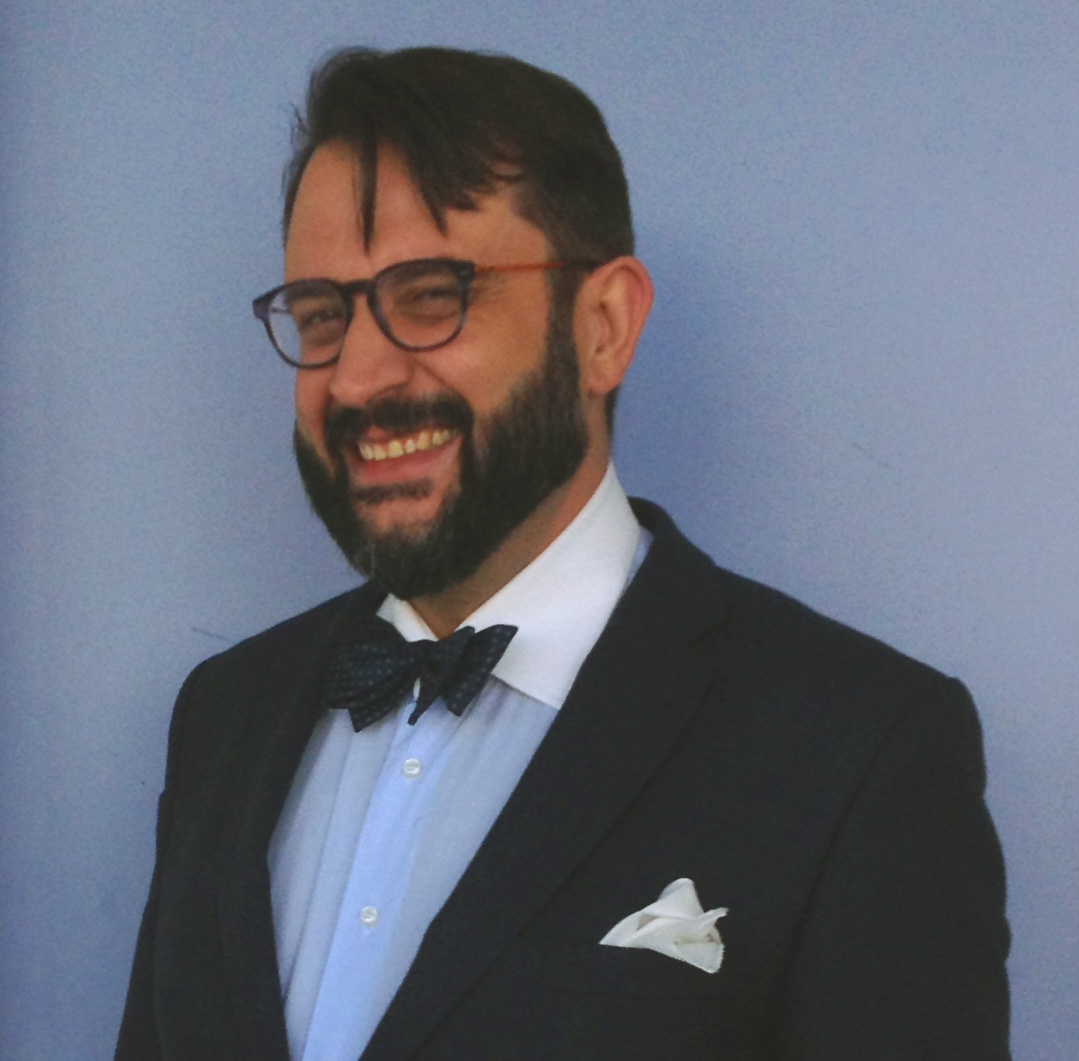}}]{Antonios Gasteratos}
received the BEng and PhD degrees from the Department of Electrical and Computer Engineering, DUTH, Greece. He is a professor and head of Department of Production and Management Engineering, Democritus University of Thrace (DUTH), Greece. He is also the director of the Laboratory of Robotics and Automation (LRA), DUTH and teaches the courses of robotics, automatic control systems, electronics, mechatronics and computer vision. During 1999–2000 he was a visiting researcher at the Laboratory of Integrated Advanced Robotics (LIRALab), DIST, University of Genoa, Italy. He has served as a reviewer for numerous scientific journals and international conferences. He is a subject editor at Electronics Letters and an associate editor at the International Journal of Optomecatronics and he has organized/coorganized several international conferences. His research interests include mechatronics and in robot vision. He has published more than 220 papers in books, journals and conferences. He is a senior member of the IEEE.
\end{IEEEbiography}

\end{document}